\documentclass[10pt,twocolumn,letterpaper]{article}

\usepackage[pagenumbers]{cvpr}

\newcommand{\cmark}{\faCheck} 

\usepackage{multirow}
\usepackage{bbding}
\usepackage{fontawesome}
\usepackage{amssymb}
\usepackage{booktabs}

\usepackage{microtype}
\definecolor{findingframe}{gray}{0.55}
\definecolor{findingfill}{gray}{0.92}
\definecolor{pendingred}{RGB}{160,48,35}

\usepackage{tcolorbox}

\newtcolorbox{findingbox}[2][]{%
  colback=gray!10,             
  colframe=gray!40,            
  coltitle=black,              
  fonttitle=\bfseries,         
  title={Finding~#2},          
  arc=1mm,                     
  boxrule=1pt,                 
  before skip=\smallskipamount,
  after skip=\smallskipamount,
  #1                           
}

\newcommand{\finding}[3]{%
  \begin{findingbox}{#1 #2}#3\end{findingbox}%
}

\newcommand{\nmark}{\textemdash}

\providecommand{\cmark}{\ensuremath{\checkmark}}
\providecommand{\nmark}{\textemdash}

\definecolor{cvprblue}{rgb}{0.21,0.49,0.74}
\usepackage[pagebackref,breaklinks,colorlinks,allcolors=cvprblue]{hyperref}

\def\confName{CVPR}
\def\confYear{2027}

\title{Not as Specialized as You Think:\\ Auditing In-Domain Retention after VLM Post-Training}

\author{
Alejandra Perez \qquad Anita Rau \qquad Chinedu Nwoye \qquad \\ Omid Mohareri \qquad Muhammad Abdullah Jamal\\[0.5em]
Intuitive Surgical Inc., Sunnyvale, USA
}

\begin{document}

\twocolumn[{%
\renewcommand\twocolumn[1][]{#1}%
\maketitle
\begin{center}
    \centering
    \captionsetup{type=figure}
    \includegraphics[width=0.95\linewidth]{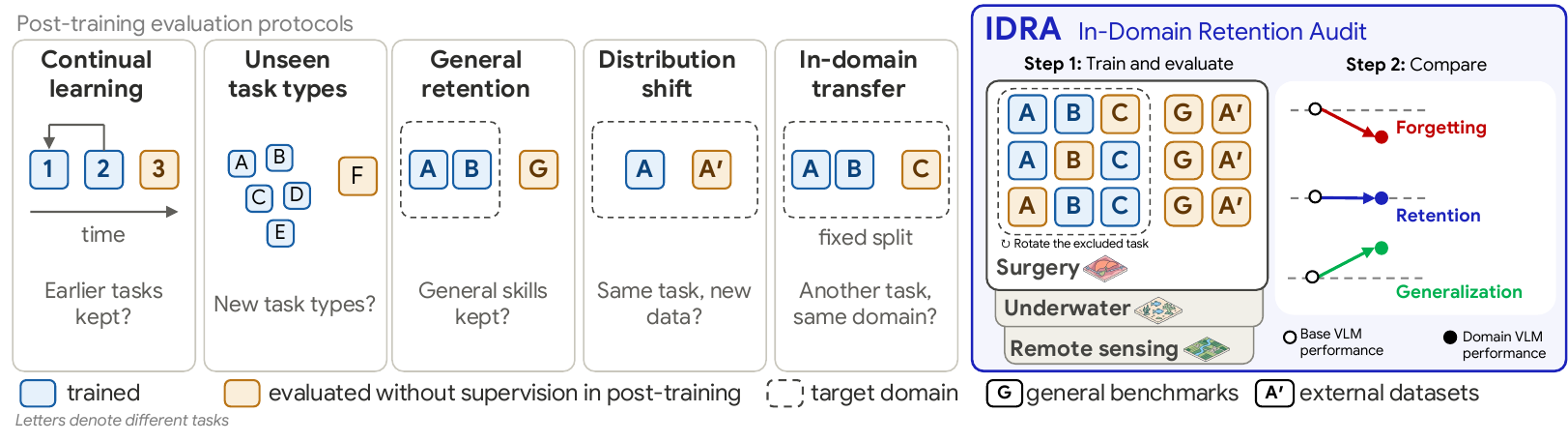}
    \captionof{figure}{\textbf{Better at the trained tasks, worse within the target domain?} Post-training can turn a general-purpose VLM into ``domain specialist'' while degrading capabilities within the domain it is meant to serve. This degradation can remain hidden behind gains on tasks supervised during post-training. \textbf{Left:} Existing evaluation protocols examine related but distinct questions; they do not collectively establish whether specialization preserves other in-domain capabilities. \textbf{Right:} Our \textbf{In-Domain Retention Audit (IDRA)} rotates the task excluded from adaptation across independent runs from the same base VLM, separating supervised acquisition from cross-task transfer. Comparing each adapted model with the base reveals whether those skills degrade, persist, or improve. Across surgery, underwater imagery, and remote sensing, IDRA surfaces hidden costs of specialization and tests whether general-purpose and external-dataset evaluations reflect them.}
    \label{fig:teaser}
\end{center}%
}]

\begin{abstract}

Post-training is widely used to leverage the broad capabilities of state-of-the-art vision-language models (VLMs) and adapt them to the specific needs of specialized domains, yet the impact of such adaptation on model generalizability within the target domain as a whole remains incompletely understood. Improving performance on the tasks used for adaptation does not establish successful domain specialization: the resulting model can become less capable than its general-purpose counterpart on new tasks within the same domain, which self-defeats the purpose of adaptation. To make this effect measurable, we introduce the \textbf{In-Domain Retention Audit (IDRA)}, a training and evaluation framework that exposes these hidden costs by assessing what adapted models retain or forget alongside what they learn. IDRA systematically excludes one task from adaptation, trains on the remaining tasks, and rotates the exclusion across independent runs, measuring changes relative to the original model. We instantiate IDRA across three visually and conceptually distinct domains -- surgery, underwater imagery, and remote sensing -- and compare supervised fine-tuning (SFT), Group Relative Policy Optimization (GRPO), and on-policy self-distillation (OPSD). Our audit surfaces in-domain skill degradation that evaluations on targeted tasks can miss. SFT can achieve large gains but also severe negative transfer, whereas GRPO generally limits degradation without consistently maximizing positive transfer. OPSD also shows mixed transfer behavior, improving some unseen tasks while substantially degrading others. By making these shortcomings measurable, IDRA establishes a practical basis for evaluating successful specialization through both capability acquisition and in-domain retention while challenging the assumption that task-specific post-training necessarily produces broader domain specialization.

\end{abstract}    
\section{Introduction}
\label{sec:intro}

The broad capabilities of vision-language models (VLMs) have motivated efforts to build on these foundations through domain-specific post-training, typically using supervised fine-tuning (SFT) or reinforcement learning (RL), with the goal of producing models that meet the needs of specialized domains~\cite{li2023llavamed,kuckreja2024geochat,xu2025nautilus,cheng2025adamllm,lai2025medr1}. Although adaptation through post-training is typically guided by available supervision within a domain; an \textit{ideal scenario} would be that skills  acquired during this stage also improve performance on other unseen tasks within the same domain -- a term known as positive cross-task transfer~\cite{wei2022flan}. Yet improved performance on the tasks used for adaptation does not necessarily transfer to other tasks within the same domain~\cite{jeong2024medical}. More importantly, post-training can even provoke an underexplored \textbf{counterintuitive effect:} the resulting “domain specialist model” may perform \textbf{worse} than its original general-purpose counterpart on tasks within the same domain that were not present in adaptation stages. Therefore, we argue that successful adaptation should improve targeted skills without degrading other in-domain capabilities. That said, the problem is that such negative cross-task transfer can remain undetected when evaluation focuses on the tasks directly targeted by adaptation or when held-out tasks remain fixed. This raises a fundamental question: \textbf{does domain-specific post-training make a VLM more capable across its target domain, or can gains on supervised tasks come at the expense of other in-domain capabilities?}

Existing evaluation protocols address related but distinct questions. Prevalent research on \textit{catastrophic forgetting} examines how learning new skills can harm existing capabilities by modifying the parameters that support them~\cite{kirkpatrick2017overcoming}. Studies on forgetting after fine-tuning primarily examine capability retention on general-purpose benchmarks or previously learned tasks~\cite{zhai2024investigating,zhu2024modeltailor,huang2025spider,zhu2026teach,chen2026retaining}. In particular, continual learning evaluates retention as new tasks are introduced sequentially~\cite{chen2024coin,zhao2025mllm,lai2026rft}, while forward transfer measures performance on later tasks before they receive supervision~\cite{lopezpaz2017gem}. Instruction tuning evaluates generalization to held-out task types, but typically relies on broad, diverse task mixtures to develop general zero-shot capabilities~\cite{wei2022flan,xu2023multiinstruct}. Domain-adapted VLMs, in turn, are commonly evaluated on supervised tasks or related domain benchmarks~\cite{li2023llavamed,kuckreja2024geochat,cheng2025adamllm,jeong2024medical}, with prior work also examining transfer across medical question types within a single benchmark~\cite{lai2025medr1}. Despite these efforts, systematic comparisons of how different post-training strategies affect unseen-task transfer within specialized visual domains remain limited. In particular, it remains unclear whether adapting a model on a subset of domain-specific tasks improves or degrades its performance on other tasks within the same domain.

To address this gap, we introduce the \textbf{In-Domain Retention Audit (IDRA)}. As shown in Figure~\ref{fig:teaser},  IDRA trains the same base model on two of three tasks and rotates the excluded task across independent runs within each domain, allowing each in-domain task to be evaluated with and without direct supervision. By measuring changes relative to the base model, we distinguish supervised acquisition from in-domain retention, quantify the frequency of degradation, and assess the cost of selecting models based on just supervised-task performance.

We apply IDRA across three visually and conceptually distinct domains such as surgery (CholecVQA~\cite{seenivasan2023surgicalgpt}), underwater imagery (NAUTILUS~\cite{xu2025nautilus}), and remote sensing (VRSBench~\cite{li2024vrsbench}), yielding nine task-exclusion splits. Within this protocol, we compare supervised fine-tuning (SFT) with two on-policy approaches: Group Relative Policy Optimization (GRPO) and on-policy self-distillation (OPSD). This comparison tests whether differences in retention reported across post-training strategies~\cite{lai2026rft,chen2026retaining,harrington2026does} extend to capabilities excluded from domain adaptation. To examine whether the observed transfer patterns extend beyond the adaptation dataset, we additionally evaluate on PhaKIR~\cite{rueckert2026video} (surgery) and HRVQA~\cite{li2024hrvqa} (remote sensing). Finally, we use MMMU~\cite{yue2024mmmu}, MathVision~\cite{wang2024measuring}, RealWorldQA~\cite{xai_realworldqa_2024}, and ODinW~\cite{li2022elevater} to test whether general-purpose retention provides useful evidence of in-domain retention.

Our results reveal a disconnect between supervised gains and in-domain retention. SFT and OPSD each produce negative mean held-out changes in five of nine splits, compared with three for GRPO, whose losses are all smaller than one metric point indicating less negative cross-task transfer than the rest but limited acquisition. 
This disconnect has consequences for model selection: choosing strategies by performance on supervised tasks yields held-out losses in four splits, including a 14.14-point accuracy drop on underwater VQA. Nor does retention on general-purpose tasks rule out these negative effects: underwater OPSD improves RealWorldQA by 3.2 accuracy points while losing 10.3 points on held-out VQA. Together, these findings show why successful domain adaptation requires evidence of retention beyond just supervised gains and general-purpose benchmark performance after post-training.

\begin{table*}[t]
\centering
\caption{\textbf{Comparison with related protocols for transfer and retention after post-training.}
\textbf{V}: generative vision-language models;
\textbf{T}: evaluation of task capabilities without direct adaptation supervision, including future tasks in sequential settings;
\textbf{D}: cross-task evaluation within a specialized application domain;
\textbf{O}: evaluation under distribution shifts across datasets, modalities, locations, or time periods within the same specialized application domain;
\textbf{G}: explicit assessment of general-purpose benchmark retention relative to the pre-adaptation model;
\textbf{R}: systematic rotation of excluded task groups, with independent training on complementary groups.
A dash denotes a feature outside the reported protocol under these definitions, not an absence of related evidence.
$\dagger$ Vero explicitly studies preservation of general visual-chat and instruction-following capabilities.}
\label{tab:protocol_comparison}
\small
\setlength{\tabcolsep}{3pt}
\renewcommand{\arraystretch}{1.12}
\resizebox{0.9\textwidth}{!}{%
\begin{tabular}{@{}lllcccccc@{}}
\toprule
\textbf{Work} & \textbf{Transfer / retention scope} & \textbf{Adaptation studied}
& \textbf{V} & \textbf{T} & \textbf{D} & \textbf{O} & \textbf{G} & \textbf{R} \\
\midrule
FLAN \cite{wei2022flan}
& Task clusters & Instruction tuning
& \nmark & \cmark & \nmark & \nmark & \nmark & \cmark \\
Tiong et al. \cite{tiong2024we}
& Vision-language tasks & Supervised fine-tuning
& \cmark & \cmark & \nmark & \nmark & \nmark & \nmark \\
Task Transfer in VLMs \cite{sachdeva2026understanding}
& Visual perception tasks & Supervised fine-tuning
& \cmark & \cmark & \nmark & \nmark & \nmark & \nmark \\
Vero \cite{sarch2026vero}
& Visual task categories / chat retention & SFT, RL
& \cmark & \cmark & \nmark & \nmark & \cmark$^{\dagger}$ & \nmark \\
Med-R1 \cite{lai2025medr1}
& Medical question types / modalities & SFT, RL
& \cmark & \cmark & \cmark & \cmark & \nmark & \nmark \\
Geo-R1 \cite{zhang2026geo}
& Geospatial referring / dataset shifts & SFT, RL
& \cmark & \nmark & \nmark & \cmark & \nmark & \nmark \\
MedVLM-R1 \cite{pan2025medvlmr1}
& Medical imaging modalities & SFT, RL
& \cmark & \nmark & \nmark & \cmark & \nmark & \nmark \\
SFT Memorizes, RL Generalizes \cite{chu2025sft}
& Rule / visual variants & SFT, RL
& \cmark & \nmark & \nmark & \nmark & \nmark & \nmark \\
\midrule
Retaining by Doing \cite{chen2026retaining}
& Non-target language capabilities & SFT, RL, on-policy data
& \nmark & \cmark & \nmark & \nmark & \cmark & \nmark \\
Self-Distillation Enables CL \cite{shenfeld2026selfdistillation}
& Prior capabilities / sequential tasks & SFT, self-distillation (SDFT)
& \nmark & \cmark & \nmark & \nmark & \cmark & \nmark \\
RFT Mitigates Forgetting \cite{lai2026rft}
& Sequential tasks / general capabilities & SFT, RL
& \cmark & \cmark & \nmark & \nmark & \cmark & \nmark \\
When Does CL Require Learning \cite{harrington2026does}
& Sequential tasks / temporal change & SFT, RL, distillation, prompting
& \nmark & \cmark & \nmark & \cmark & \nmark & \nmark \\
\midrule
\textbf{IDRA (ours)}
& \textbf{Excluded in-domain tasks} & \textbf{SFT, RL, OPSD}
& \cmark & \cmark & \cmark & \cmark & \cmark & \cmark \\
\bottomrule
\end{tabular}%
}
\end{table*}

Our contributions are threefold:
\begin{itemize}\setlength\itemsep{1pt}
\item \textbf{A retention-centered evaluation framework across specialized domains.} We introduce IDRA, a systematic evaluation protocol that distinguishes supervised-task acquisition from unseen in-domain capability retention across nine task-exclusion splits spanning three specialized visual domains.

\item \textbf{Characterizing negative transfer under post-training.} Across SFT, GRPO, and OPSD, we show that strong supervised-task gains do not guarantee positive transfer to unseen in-domain tasks. We further demonstrate that supervised-task rankings can reverse on held-out tasks and that model selection based on supervised performance can favor configurations that degrade unseen capabilities within the target domain.

\item \textbf{Evaluating generalization beyond supervised tasks.} We show that improvements on source datasets do not necessarily generalize to external datasets within the same domain, and that preserving general-purpose capabilities does not establish retention of other in-domain skills.

\end{itemize}

\section{Related Work}
\label{sec:related_work}

\noindent \textbf{Domain-adapted VLMs.} Generalist VLMs are routinely specialized to biomedicine~\cite{li2023llavamed}, remote sensing~\cite{kuckreja2024geochat}, egocentric views~\cite{ye2025mmego}, and surgery~\cite{wang2026endochat}, through SFT and increasingly RL~\cite{pan2025medvlmr1, lai2025medr1, zhang2026geo, li2026coa}. Evaluation commonly focuses on supervised tasks or new datasets of those tasks, where RL-based adaptation can transfer better than SFT~\cite{pan2025medvlmr1, zhang2026geo, li2026coa}. However, domain post-training can be on par or even fall below the original model on specialized benchmarks~\cite{jeong2024medical}.
Med-R1~\cite{lai2025medr1} is particularly relevant because it evaluates transfer across medical question types and imaging modalities. We study retention and acquisition under rotating task exclusion across several domains and post-training strategies, including tasks with different prediction targets and structures.

\noindent \textbf{Forgetting under SFT, RL, and OPSD.} Fine-tuning on a target task can degrade capabilities outside it~\cite{zhai2024investigating, zhu2024modeltailor, huang2025spider, biderman2024lora}. Several studies, mostly on LLMs, find that RL with GRPO~\cite{shao2024deepseekmath} generalizes better and forgets less than SFT~\cite{chu2025sft, huan2025math, lai2026rft}, attributing this to implicit regularization~\cite{lai2026rft}, a preference for solutions close to the base model~\cite{shenfeld2025rlsrazor}, on-policy data~\cite{chen2026retaining}, or limited output-distribution shifts~\cite{zhu2026teach}. On-policy distillation~\cite{agarwal2024policy} combines on-policy data with dense token-level supervision. Self-distillation variants use the model itself as teacher, conditioned on reference solutions~\cite{zhao2026opsd}, demonstrations~\cite{shenfeld2026selfdistillation}, environment feedback~\cite{hubotter2026reinforcement}, or alternative image views~\cite{yuan2026visionopd, li2026s2vopd, aniri2026opdv}. Retention results are mixed: SDFT forgets less than SFT~\cite{shenfeld2026selfdistillation}, whereas self-distillation policy optimization (SDPO) forgets more than GRPO in continual post-training~\cite{wang2026denser}. These comparisons primarily assess general retention, earlier tasks, or trained-task variants. Retention on these evaluations does not establish retention of excluded capabilities within the target domain.

\noindent \textbf{Evaluation protocols for task transfer.} Table~\ref{tab:protocol_comparison} compares related protocols. Continual learning measures backward and forward transfer along task sequences~\cite{lopezpaz2017gem, chen2024coin, zhao2025mllm}, including adaptation across domains and temporal changes~\cite{harrington2026does}. Instruction tuning evaluates held-out task clusters to study generalist zero-shot ability~\cite{wei2022flan, xu2023multiinstruct}. Taskonomy and subsequent multitask studies map relationships between tasks~\cite{zamir2018taskonomy, standley2020tasks}. More directly, Tiong et al.~\cite{tiong2024we} and Sachdeva et al.~\cite{sachdeva2026understanding} study how VLM fine-tuning affects other tasks, including positive and negative transfer. Vero~\cite{sarch2026vero} examines transfer across visual task categories and preservation of general visual capabilities. These studies establish that task gains need not transfer uniformly which motivates a post-training study where trained and held-out tasks are not fixed.

\section{Preliminaries}
\label{sec:prelim}

\noindent\textbf{Domain adaptation setting.}
A VLM defines an autoregressive policy $\pi_\theta(y\mid x)$ over responses $y$ to an image--instruction input $x$. Adaptation updates a base policy $\pi_{\theta_0}$ using data $\mathcal D$ from tasks $\mathcal S_d\subset\mathcal T_d$ in domain $d$. We assess acquisition on $\mathcal S_d$ and retention on $\mathcal T_d\setminus\mathcal S_d$. ``Base'' denotes the starting checkpoint; ``held out'' refers only to the current adaptation stage.

\noindent\textbf{Supervised fine-tuning.}
SFT maximizes reference-response likelihood:
\begin{equation}
\mathcal{L}_{\mathrm{SFT}}(\theta)
=
-\mathbb{E}_{(x,y^*)\sim\mathcal{D}}
\left[
\sum_{t=1}^{|y^*|}
\log\pi_\theta(y_t^*\mid x,y_{<t}^*)
\right].
\label{eq:sft}
\end{equation}
Conditioning on reference rather than generated prefixes creates a train--inference mismatch~\cite{harrington2026does}, while fitting reference responses can impair non-target capabilities~\cite{chen2026retaining}.

\noindent\textbf{Reinforcement learning with verifiable rewards.}
RL maximizes task-specific rewards on generated responses:
\begin{equation}
J_{\mathrm{RL}}(\theta)
=
\mathbb{E}_{x\sim\mathcal{D},\,y\sim\pi_\theta(\cdot\mid x)}
[r(x,y)].
\label{eq:rl}
\end{equation}
We use GRPO~\cite{shao2024deepseekmath}, which samples $G$ responses per input and applies a clipped policy-ratio objective with centered advantages:
\begin{equation}
\hat A_i=r(x,y_i)-\frac{1}{G}\sum_{j=1}^{G}r(x,y_j).
\label{eq:advantage}
\end{equation}
Prior work links on-policy learning and selective updates to reduced forgetting~\cite{chen2026retaining,lai2026rft}.

\noindent\textbf{On-policy self-distillation.}
OPSD~\cite{zhao2026opsd} supervises student-generated responses using a frozen base teacher ($\bar\theta=\theta_0$) given the reference answer $y^*$. At each generated prefix,
\begin{equation}
\begin{aligned}
p_{\theta,t} &= \pi_\theta(\cdot\mid x,y_{<t}),\\
q_{\bar\theta,t} &= \pi_{\bar\theta}(\cdot\mid x,y^*,y_{<t}).
\end{aligned}
\end{equation}
The student minimizes reverse KL along its trajectories:
\begin{equation}
\mathcal{L}_{\mathrm{OPSD}}(\theta)
=
\mathbb{E}_{\substack{(x,y^*)\sim\mathcal{D}\\
y\sim\pi_\theta(\cdot\mid x)}}
\left[
\frac{1}{|y|}
\sum_{t=1}^{|y|}
D_{\mathrm{KL}}\!\left(p_{\theta,t}\,\|\,q_{\bar\theta,t}\right)
\right].
\label{eq:opsd}
\end{equation}
GRPO and OPSD thus both learn on generated trajectories, using response-level rewards and token-level distributions, respectively.
\section{In-Domain Retention Audit}
\label{sec:protocol}

\begin{figure*}[t]
\centering
\includegraphics[width=0.95\linewidth]{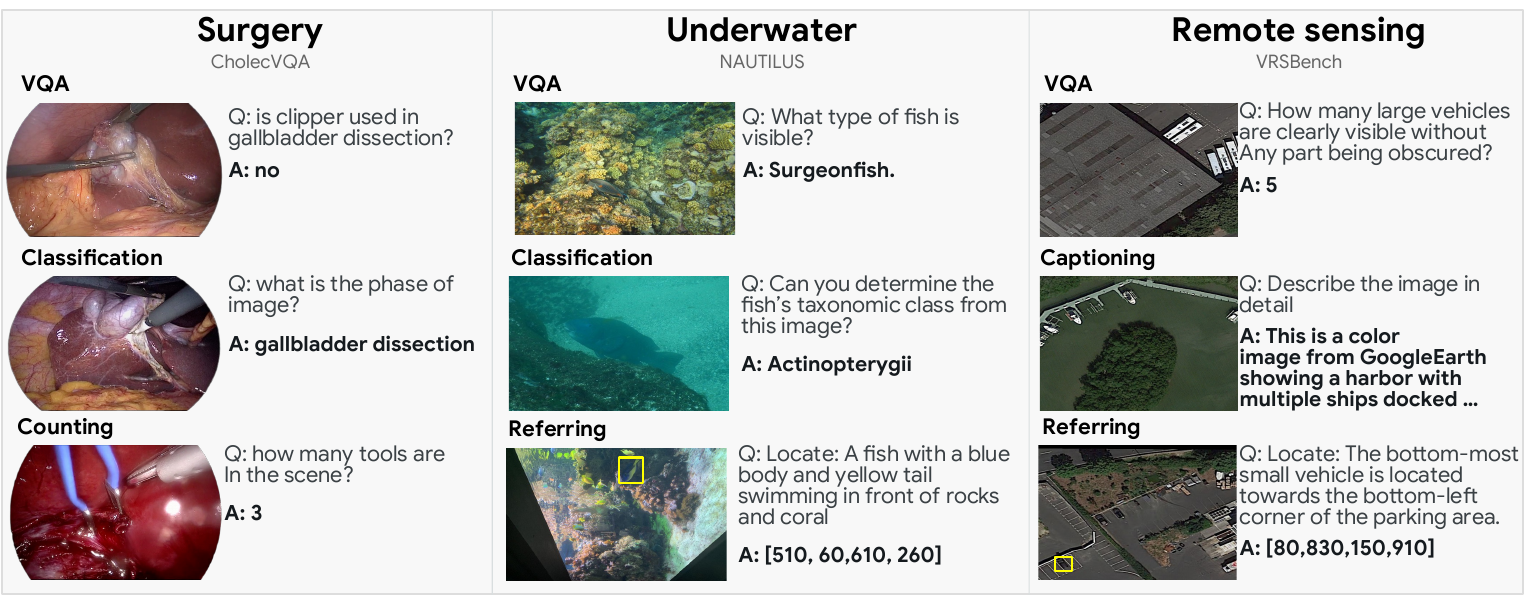}
\caption{\textbf{Representative examples of the domains and tasks.} Examples illustrate the specialized imagery, prediction targets, and output formats in each domain for surgery, underwater scene understanding and remote sensing. }
\label{fig:overview}
\end{figure*}

IDRA separates three questions: what adaptation acquires under supervision, what it preserves on excluded tasks, and whether either outcome extends beyond the source dataset. General benchmarks provide a fourth, complementary assessment whose agreement with in-domain retention is explicitly tested.

\subsection{Rotating task exclusion}
\label{sec:protocol_loto}
For each domain, we rotate the excluded task $h_s$ among $\mathcal T_d=\{T_1,T_2,T_3\}$, independently training on the other two. This evaluates each capability with and without direct supervision. All methods start from the same base model, use identical training examples within each split, and select checkpoints using supervised-task validation alone. 

\noindent\textbf{Domains and supervision.}
IDRA requires specialized visual datasets with supervision for at least three vision-language tasks and both training data and a separate benchmark test set. We select CholecVQA~\cite{seenivasan2023surgicalgpt}, derived from Cholec80~\cite{twinanda2016endonet}, NAUTILUS~\cite{xu2025nautilus}, and VRSBench~\cite{li2024vrsbench} to meet these requirements (\cref{fig:overview,tab:datasets}). To control supervision volume, we sample 3,000 training examples per task for surgery and 5,000 for underwater imagery and remote sensing. Classification labels are balanced for training where feasible and test distributions remain unchanged.

\noindent For external and general evaluations, we also train each method on all three tasks (\emph{All}) as a full-supervision reference.

\begin{table}[t]
\centering\small
\setlength{\tabcolsep}{4pt}
\begin{tabular}{@{}lll@{}}
\toprule
Domain / dataset & Task & Metric\\
\midrule
\multirow{3}{*}{Surgery / CholecVQA} & Binary VQA & Macro-F1\\
& Phase classification & Macro-F1\\
& Counting & Macro-F1\\
\midrule
\multirow{3}{*}{Underwater / NAUTILUS} & Classification & Accuracy\\
& VQA & Accuracy\\
& Referring & Acc@0.5\\
\midrule
\multirow{3}{*}{Remote Sensing / VRSBench} & VQA & Accuracy\\
& Referring & Acc@0.5\\
& Captioning & CLAIR\\
\bottomrule
\end{tabular}
\caption{\textbf{Tasks and metrics.} Each task uses its native metric. Binary VQA and counting (0--3 tools) use macro-F1 so that each answer class has equal weight. Acc@0.5 is the fraction of boxes with IoU${\geq}0.5$; captioning uses CLAIR~\cite{chan2023clair}.}
\label{tab:datasets}
\end{table}

\subsection{Acquisition and retention measures}
\label{sec:protocol_metrics}
For method $m$, split $s$, and task $T$, we compare the adapted policy with the base under the same evaluation prompt $c_m$:
\begin{equation}
\Delta_{m,s,T}
=
f_T(\pi_{m,s};c_m)-f_T(\pi_{\theta_0};c_m),
\label{eq:delta}
\end{equation}
where $f_T$ is the task score. Changes on supervised tasks measure acquisition; $\Delta_{m,s,h_s}$ measures excluded-task retention. We report tasks separately in their native metrics to avoid masking losses through aggregation. \\

\noindent\textbf{Held-out degradation rate.}
HDR measures the fraction of splits whose mean held-out loss exceeds a task-specific tolerance:
\begin{equation}
\operatorname{HDR}_m(\boldsymbol\epsilon)
=
\frac1S\sum_{s=1}^{S}
\mathbf{1}[\overline\Delta_{m,s,h_s}<-\epsilon_{h_s}],
\label{eq:hdr}
\end{equation}
where $S=9$, $\overline\Delta$ averages changes across runs, and $\epsilon_{h_s}\geq0$. We use $\boldsymbol\epsilon=0$ to count negative means and examine sensitivity to other tolerances. Counting events avoids averaging magnitudes across heterogeneous metrics. Per-task changes with 95\% intervals, tolerance sensitivity, absolute scores, and the configurations selected under Eq.~\ref{eq:selection_cost} are in \cref{supp:results}.\\

\begin{figure*}[t]
\centering
\includegraphics[width=0.95\linewidth]{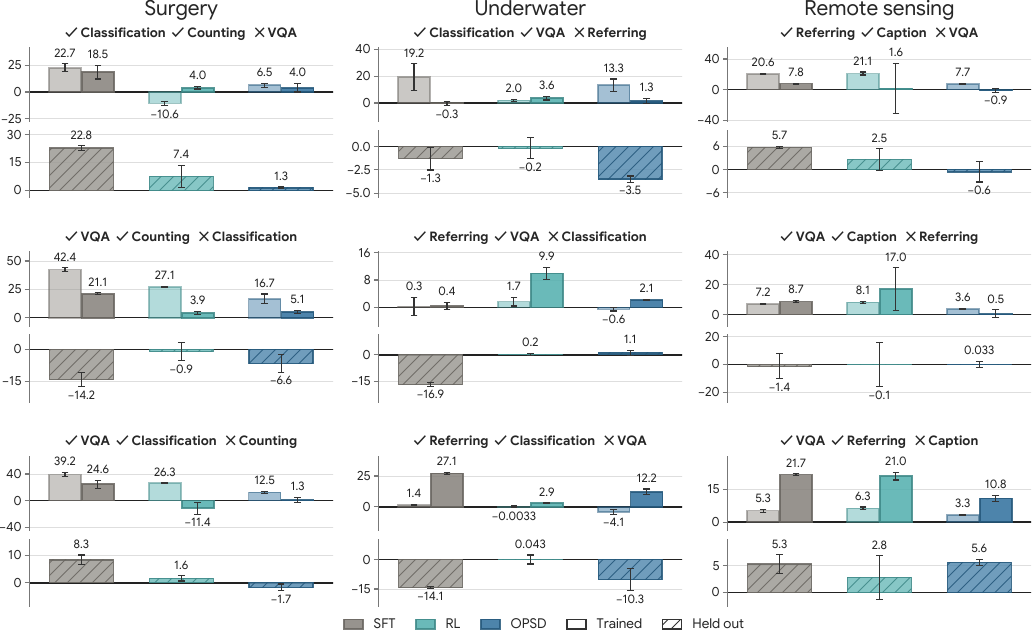}
\caption{\textbf{Supervised gains can coexist with losses on excluded in-domain tasks.} Across nine task-exclusion splits, solid bars show changes on the two supervised tasks (in subtitle order), while hatched bars show changes on the excluded task, relative to the base model. SFT yields both the largest held-out gain and loss while GRPO limits held-out degradation but can fail on supervised tasks. Each band uses its own scale and task-specific metric (\cref{tab:datasets}). Error bars show standard deviation across three training runs.}
\label{fig:main_results}
\end{figure*}

\noindent\textbf{Cost of supervised selection.}
Within each run, checkpoints maximize the mean validation score of the two supervised tasks. Averaging this objective across runs gives $V_s(m)$; we retrospectively select $\widehat m_s=\arg\max_m V_s(m)$ and measure its held-out loss relative to the base:
\begin{equation}
C_s=\max\{0,-\overline\Delta_{\widehat m_s,s,h_s}\}.
\label{eq:selection_cost}
\end{equation}
We report costs per task and count selected configurations with negative held-out changes. The supplement provides selected methods, absolute scores, and costs.

\subsection{External transfer and general retention}
\label{sec:protocol_external}
We test external in-domain transfer on PhaKIR phase classification~\cite{rueckert2026video} and HRVQA remote-sensing VQA~\cite{li2024hrvqa}, distinguishing whether the corresponding source task was supervised or excluded. To assess agreement between general and in-domain retention, we compare held-out changes with accuracy changes on MMMU~\cite{yue2024mmmu}, MathVision~\cite{wang2024measuring}, and RealWorldQA~\cite{xai_realworldqa_2024}, and mean COCO AP changes across 12 ODinW-13 datasets~\cite{li2022elevater}.

\subsection{Training and evaluation controls}
\label{sec:protocol_strategies}
 All methods adapt Qwen3.5-4B with thinking mode disabled, updating all parameters for one epoch with the same training examples, batch size, and learning rate, so all configurations within a split take the same number of steps. Hyperparameters, prompts, rewards, and the extractor are detailed in \cref{supp:prompts,supp:training,supp:eval}.

\noindent\textbf{Prompting.} SFT is trained and evaluated with a prompt that poses the question without asking for visual reasoning traces, since reference annotations do not provide them. GRPO and OPSD use a describe-then-answer prompt, $y{=}(z,\hat a)$, except for captioning. On-policy methods learn from what the student samples: GRPO obtains a nonzero advantage only when sampled answers differ in reward, and the OPSD teacher can correct intermediate visual interpretations only if they are generated. Because prompts shift base performance, each adapted model is compared with the base under its own prompt (Eq.~\ref{eq:delta}), which accounts for different starting points.

\noindent\textbf{Answer extraction.} Before scoring, we recover the task answer with deterministic parsing and, if parsing fails we use an LLM extractor (\eg, box coordinates or a class name embedded in explanatory text). Therefore, metrics are applied to the recovered answer, so scores reflect task capability and not output format. 
\section{Results}
\label{sec:results}

\subsection{Supervised rankings can reverse on excluded tasks}
\label{sec:results_selection}
\Cref{fig:main_results} provides a direct ranking reversal. When surgical phase classification is held out, SFT achieves the largest gains on both supervised tasks: +42.4 macro-F1 on VQA and +21.1 on counting, compared with +27.1/+3.9 for GRPO and +16.7/+5.1 for OPSD. Yet SFT loses 14.2 points on the excluded task, compared with losses of 0.9 and 6.6 for GRPO and OPSD. The configuration that leads on each supervised gain therefore has the worst held-out change. 

Large losses also occur without substantial supervised gains. Underwater SFT loses 16.9 points on held-out classification while improving supervised referring and VQA by only 0.3 and 0.4 points. Conversely, surgical SFT improves both supervised tasks by approximately 20 points while gaining 22.8 points on held-out VQA. These configurations show why neither strong acquisition nor weak acquisition determines the direction of held-out change. 

\noindent\textbf{How often do excluded capabilities degrade?}
Across the nine task-exclusion splits, HDR at zero tolerance is 55.6\% for SFT and OPSD, compared with 33.3\% for GRPO (\cref{tab:hdr}). However, the three negative GRPO means are all smaller than one point in magnitude. When we allow a tolerance of one native metric point per task, GRPO's HDR falls to zero, while SFT remains at 55.6\% and OPSD falls to 44.4\%. This distinction shows that GRPO's negative changes are concentrated near zero, whereas the losses under SFT and OPSD persist beyond this tolerance. Equal HDR values nevertheless need not imply similar outcomes: on held-out underwater classification, SFT loses 16.9 accuracy points while OPSD gains 1.1. We therefore interpret HDR alongside the individual task changes.

\noindent\textbf{What is the cost of selecting from supervised scores alone?}
We next apply the supervised-validation selection rule defined in \cref{eq:selection_cost}. The selected configuration falls below its prompt-matched base in four of nine splits, yielding costs of 14.20 macro-F1 points on surgical phase recognition, 3.5 Acc@0.5 points on underwater referring, 14.14 accuracy points on underwater VQA, and 0.07 Acc@0.5 points on remote referring. The remaining five splits have zero cost because their selected configurations match or improve on the base. Although the remote-referring change is near zero, the larger losses show that selecting by supervised validation alone can favor configurations that degrade excluded capabilities. 

\finding{1}{}{Supervised gains leave retention unresolved. A ranking based on acquisition can favor the configuration with the greatest held-out degradation.}

\begin{table}[t]
\centering\small
\begin{tabular}{@{}lccc@{}}
\toprule
Negative held-out mean changes & SFT & GRPO & OPSD\\\midrule
Surgery & 1/3 & 1/3 & 2/3\\
Underwater & 3/3 & 1/3 & 2/3\\
Remote sensing & 1/3 & 1/3 & 1/3\\\midrule
All holdouts & 5/9 & 3/9 & 5/9\\
HDR at $\boldsymbol\epsilon=0$ & 55.6\% & 33.3\% & 55.6\%\\\bottomrule
\end{tabular}
\caption{\textbf{GRPO's held-out losses are less frequent and smaller than those of SFT and OPSD.} Counts indicate negative mean changes across the nine task-exclusion splits, not statistical significance. Allowing a one-native-point loss reduces degradation counts from 5/9, 3/9, and 5/9 to 5/9, 0/9, and 4/9 for SFT, GRPO, and OPSD, respectively. }
\label{tab:hdr}
\end{table}

\subsection{Retention depends on the capability and supervision}
\label{sec:results_tasks}
\noindent\textbf{Direct supervision changes the classification outcome.}
Under SFT, surgical phase classification gains 22.7 and 24.6 macro-F1 points when supervised, but loses 14.2 when excluded. Underwater classification similarly gains 19.2 and 27.1 accuracy points with supervision, but loses 16.9 without it. Thus, success at acquiring a capability does not imply that training on neighboring tasks preserves it. Because each run jointly supervises two tasks, these contrasts identify dependence on the supervision mixture rather than the isolated effect of either source task.

The surgical asymmetry between phase classification and VQA follows from how CholecVQA uses phases. Every binary question names a phase (\eg, ``is grasper used in calot triangle dissection?''), so the phase is given rather than inferred from the image. Phase supervision thus provides the scene context these questions rely on, whereas VQA supervision exposes phase names without requiring phase recognition. Without phase supervision, SFT does not just fail to improve but collapses toward a single phase: gallbladder dissection receives most predictions. All outputs remain valid phase names, so the loss lies in mapping images to phases rather than in output format (\cref{supp:surgery_transfer}).

Rotation also changes the conclusion a fixed holdout would support. If only VQA or counting were held out in surgery, SFT would appear to transfer positively (+22.8 and +8.3); rotating the holdout reveals a 14.2-point loss on phase classification. Similarly, held-out captioning improves under all three strategies, whereas held-out referring changes range from $-3.5$ to $+0.03$ Acc@0.5. Conclusions from a single fixed holdout can therefore depend on which task happens to be excluded.

\finding{2}{}{Rotation is necessary. The same adaptation can help one excluded task and hurt another, so a single fixed holdout can support the opposite conclusion.}

\noindent\textbf{GRPO's retention does not imply uniform acquisition.}
GRPO has positive mean changes on six held-out tasks and small negative means on the other three. This preservation coexists with substantial supervised gains on surgical VQA (+27.1/+26.3) and remote referring (+21.1/+21.0), so it cannot be explained uniformly by an absence of learning. However, supervised surgical phase classification falls by 10.6 and 11.4 macro-F1 points, compared with only 0.9 when held out. 

Rollout statistics locate this failure. Phase labels are imbalanced, and for rare phases such as clipping/cutting nearly all eight sampled answers were wrong, leaving zero advantage and no corrective signal. GRPO therefore collapsed onto frequent classes: rare-phase F1 fell to near zero while the frequent packaging class improved (\cref{supp:grpo_phase}).

\noindent\textbf{On-policy sampling does not establish retention.}
OPSD improves supervised underwater classification by 13.3 and 12.2 points, but loses 10.3 points on held-out underwater VQA and 6.6 on surgical phase classification. Its retention profile therefore differs substantially from GRPO despite both using student-generated trajectories. 

\finding{3}{}{Retention and acquisition must be assessed together: GRPO can preserve excluded tasks while failing supervised ones, and OPSD can do the opposite despite also learning on-policy.}

\subsection{Source gains can reverse on external domain data}
\label{sec:results_external}
\begin{figure}[t]
\centering
\includegraphics[width=\linewidth]{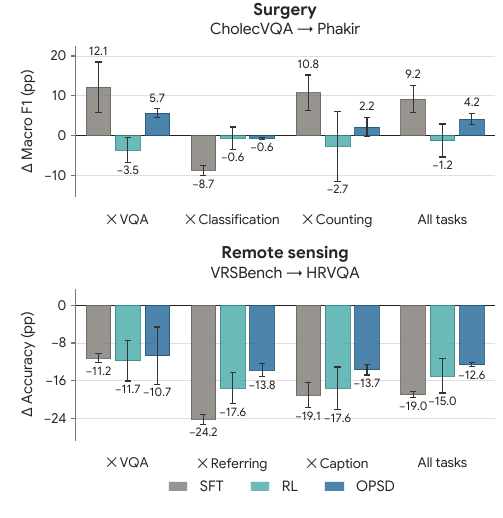}
\caption{\textbf{Source-dataset gains do not ensure transfer to external in-domain data.} Changes relative to base model on PhaKIR phase classification and HRVQA. SFT improves on PhaKIR only when phase classification receives supervision, while all twelve configurations degrade on HRVQA, including those with source-dataset VQA gains. Crosses identify the task excluded from adaptation and ``All tasks'' denotes complete source-task supervision. }
\label{fig:ood_transfer}
\end{figure}

On PhaKIR, the dependence on direct phase supervision persists (\cref{fig:ood_transfer}). SFT gains 9.2--12.1 macro-F1 points when phase classification is supervised, but loses 8.7 when it is excluded. GRPO loses 0.6--3.5 points across all four configurations, consistent with its failure to improve phase classification on the source dataset. This correspondence shows that the source pattern can extend to a different surgical dataset.

HRVQA exposes a different limitation: all remote-sensing configurations degrade, including those that improve VRSBench VQA. Unlike VRSBench, whose questions almost always refer to objects present in the image, HRVQA often asks about absent ones (\eg, ``How large is the soccer field?'' when there is none). Most of the degradation comes from these questions (\cref{supp:hrvqa}), suggesting that adapting to VRSBench biases models toward assuming the queried object exists.

\subsection{General benchmarks provide an inconsistent in-domain retention signal}
\label{sec:results_general}
\begin{figure}[t]
\centering
\includegraphics[width=\linewidth]{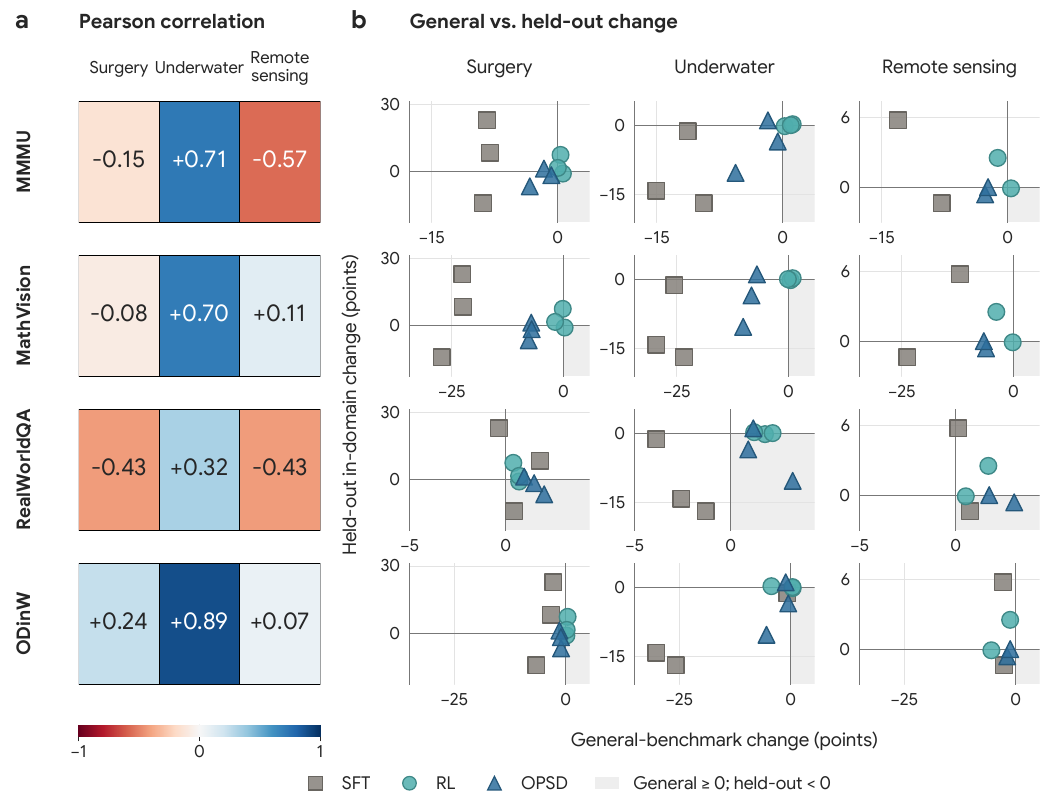}
\caption{\textbf{General-benchmark retention is not a reliable proxy for in-domain retention.} (a) Pearson correlations between general-benchmark and held-out task changes vary across domains, with positive associations in underwater imagery but weak or negative associations in most surgery and remote-sensing comparisons. (b) Changes relative to the base model; shaded regions mark preserved or improved general performance alongside in-domain degradation. Each correlation uses $n=9$ configurations and is descriptive.}

\label{fig:general_vs_indomain}
\end{figure}

General-retention profiles depend on the strategy and benchmark. Across all configurations, SFT loses 7.9--15.1 accuracy points on MMMU and 12.0--29.6 on MathVision. GRPO stays within 1.2 points of the base on MMMU and 3.8 on MathVision, but loses up to 11.6 mAP points on ODinW. OPSD changes RealWorldQA by $-0.2$ to $+3.2$ points while losing 6.1--10.1 points on MathVision (\cref{fig:supp_general}).

Comparing these changes with held-out changes \cref{fig:main_results} shows that general scores can disagree with changes on the excluded task. Underwater OPSD gains 3.2 accuracy points on RealWorldQA while losing 10.3 on held-out VQA; remote-sensing OPSD similarly gains 3.1 on RealWorldQA while losing 0.6 on held-out VQA. The disagreement also occurs in the other direction: surgical SFT loses 22.6 points on MathVision while gaining 22.8 macro-F1 on held-out VQA. Improvement on an individual general benchmark therefore does not establish in-domain preservation, and general degradation does not establish in-domain loss, highlighting the need for in-domain forgetting frameworks such as IDRA.

Across 27 configurations, nonnegative general changes coexist with negative held-out changes in 4 cases for MMMU, 2 for MathVision, 2 for ODinW, and 10 for RealWorldQA (\cref{supp:general}). For RealWorldQA, the paired-bootstrap count range is 7--13, conditional on fixed general scores and checkpoints. In remote sensing, Pearson $r$ is $-0.57$ for MMMU, $+0.11$ for MathVision, $-0.43$ for RealWorldQA, and $+0.07$ for ODinW (\cref{fig:general_vs_indomain}).

\finding{4}{}{General benchmarks supply complementary evidence, but general retention is not a proxy for in-domain retention.}

\section{Conclusions}
\label{sec:conclusion}
In this work, we introduce the In-Domain Retention Audit (IDRA), a protocol for evaluating supervised acquisition and excluded-task retention across surgery, underwater imagery, and remote sensing. Our experiments reveal three shortcomings in how successful specialization is assessed. First, supervised gains can coexist with substantial excluded-task losses: SFT improves surgical VQA and counting by 42.4 and 21.1 macro-F1 points while degrading excluded phase recognition by 14.2 points. Second, retention alone does not establish successful adaptation: GRPO generally limits excluded-task degradation but can does not acquire supervised capabilities at the same rate as SFT does. Third, improvements on source datasets or individual general-purpose benchmarks do not establish preservation of other skills within the target domain. Together, these findings show that successful specialization requires evaluating both what a model learns and what it loses within its intended domain. IDRA provides a practical foundation for assessing these changes and build better domain-specific models.

\noindent \textbf{Limitations} Observed transfer patterns in this study may vary with model scale, task composition, and supervision volume. Although methods share training examples and budgets, their optimization dynamics differ. In addition, the selected tasks and external benchmarks sample only part of each domain: preserving these capabilities does not establish domain-wide retention.

{
    \small
    \bibliographystyle{ieeenat_fullname}
    \bibliography{main}
}

\clearpage
\setcounter{page}{1}
\maketitlesupplementary

\setcounter{topnumber}{4}
\setcounter{bottomnumber}{3}
\setcounter{totalnumber}{6}
\setcounter{dbltopnumber}{3}
\renewcommand{\topfraction}{0.95}
\renewcommand{\bottomfraction}{0.90}
\renewcommand{\textfraction}{0.05}
\renewcommand{\floatpagefraction}{0.75}
\renewcommand{\dbltopfraction}{0.95}
\renewcommand{\dblfloatpagefraction}{0.75}
\setlength{\floatsep}{8pt plus 2pt minus 2pt}
\setlength{\textfloatsep}{10pt plus 2pt minus 2pt}
\setlength{\dblfloatsep}{8pt plus 2pt minus 2pt}
\setlength{\dbltextfloatsep}{10pt plus 2pt minus 2pt}

\newsavebox{\suppcompactbox}
\newenvironment{suppcompacttable}{\begin{lrbox}{\suppcompactbox}}{%
  \end{lrbox}%
  \ifdim\wd\suppcompactbox>\linewidth
    \resizebox{\linewidth}{!}{\usebox{\suppcompactbox}}%
  \else\usebox{\suppcompactbox}\fi}
\newcommand{\suppscore}[2]{\shortstack[r]{#1\\(#2)}}

\definecolor{supppromptgray}{gray}{0.96}
\definecolor{supppromptedge}{gray}{0.81}
\newbox\suppprompttext
\newbox\supppromptline

\newcommand{\supppromptrow}[3]{%
  \nointerlineskip\hbox to \linewidth{%
    \rlap{\color{supppromptgray}\vrule width\linewidth height#1 depth#2}%
    {\color{supppromptedge}\vrule width.35pt}%
    \kern5pt#3\hfil{\color{supppromptedge}\vrule width.35pt}}}
\newcommand{\supppromptrule}{%
  \nointerlineskip\hbox{\color{supppromptedge}\vrule width\linewidth height.35pt depth0pt}}
\newenvironment{suppprompt}{%
  \par\nobreak\vspace{2pt}\begingroup
  \footnotesize\ttfamily
  \setbox\suppprompttext=\vbox\bgroup
    \hsize=\dimexpr\linewidth-10.7pt\relax
    \raggedright\parindent=0pt
    \interlinepenalty=0 \clubpenalty=0 \widowpenalty=0
    \noindent\ignorespaces
}{%
    \par\egroup
  \splittopskip=0pt \splitmaxdepth=\maxdimen \vbadness=10000 \vfuzz=\maxdimen
  \supppromptrule\nobreak\supppromptrow{3pt}{0pt}{}\nobreak
  \loop\ifdim\ht\suppprompttext>0pt
    \setbox\supppromptline=\vsplit\suppprompttext to 0pt
    \setbox\supppromptline=\vbox{\unvbox\supppromptline}%
    \supppromptrow{.76\baselineskip}{.24\baselineskip}{\color{black}\box\supppromptline}%
    \ifdim\ht\suppprompttext>0pt \penalty0 \fi
  \repeat
  \nobreak\supppromptrow{3pt}{0pt}{}\nobreak\supppromptrule
  \endgroup\par\addvspace{5pt}%
}

\setcounter{section}{0}\renewcommand{\thesection}{S\arabic{section}}
\setcounter{table}{0}\renewcommand{\thetable}{S\arabic{table}}\renewcommand{\theHtable}{S\arabic{table}}
\setcounter{figure}{0}\renewcommand{\thefigure}{S\arabic{figure}}\renewcommand{\theHfigure}{S\arabic{figure}}
\setcounter{equation}{0}\renewcommand{\theequation}{S\arabic{equation}}\renewcommand{\theHequation}{S\arabic{equation}}
\renewcommand{\theHsection}{S\arabic{section}}

\noindent \textbf{Overview.}
This supplement provides the dataset construction, prompting, training, and evaluation details underlying IDRA (Secs.~\ref{supp:data}--\ref{supp:eval}), complete acquisition and retention results (Sec.~\ref{supp:results}), and additional analyses supporting the main findings (Sec.~\ref{supp:analyses}). We also examine surgical transfer from a second base model with weaker starting capabilities (Sec.~\ref{supp:arch}).

\begin{table}[!htbp]
\centering
\setlength{\abovecaptionskip}{3pt}
\setlength{\belowcaptionskip}{3pt}
\footnotesize
\setlength{\tabcolsep}{3pt}
\renewcommand{\arraystretch}{1.0}
\caption{Guide to the supplementary evidence.}
\label{tab:supp_guide}
\begin{suppcompacttable}
\begin{tabular}{@{}lll@{}}
\toprule
Main paper & Question & Supplement\\\midrule
Sec. 4, Table 2 & Data, training, and scoring & \ref{supp:data}--\ref{supp:eval}\\
Sec. 4, Table 2 & Task prompts and judge templates & \ref{supp:prompts}, \ref{supp:judge_templates}\\
Fig. 3, Table 3 & Acquisition and retention & \ref{supp:results}\\
Eq. 8 & Supervised selection & \ref{supp:results}\\
Sec. 5.2 & Surgical phase failures & \ref{supp:grpo_phase}--\ref{supp:surgery_transfer}\\
Fig. 4 & External-transfer losses & \ref{supp:external}\\
Fig. 5 & General-retention disagreement & \ref{supp:general}\\
Starting capability & Surgical transfer from a weaker base & \ref{supp:arch}\\
\bottomrule\end{tabular}
\end{suppcompacttable}
\end{table}

\section{Datasets and task construction}\label{supp:data}
\subsection{Source datasets}
Table~\ref{tab:supp_data} reports the number of annotations used for each task. We fix the sampled training pool using seed 42 and reuse it across methods and training repeats. Each task-exclusion split retains only the two supervised tasks from this pool. A surgical held-out rotation therefore uses 6,000 annotations, while the separately trained all-task reference uses 9,000.

\begin{table}[!htbp]
\centering
\setlength{\abovecaptionskip}{3pt}
\setlength{\belowcaptionskip}{3pt}
\footnotesize
\setlength{\tabcolsep}{3pt}
\renewcommand{\arraystretch}{1.0}
\caption{\noindent \textbf{Source-task split sizes.} Counts denote annotations and training pools are reused across methods and repeats.}
\label{tab:supp_data}
\begin{suppcompacttable}
\begin{tabular}{@{}llrrr@{}}
\toprule
Domain & Task & Train & Val & Test\\\midrule
Surgery & Binary VQA & 3000 & 200 & 7034\\
Surgery & Phase & 3000 & 200 & 1044\\
Surgery & Counting & 3000 & 200 & 1018\\
Underwater & Classif. & 5000 & 200 & 990\\
Underwater & VQA & 5000 & 200 & 1422\\
Underwater & Referring & 5000 & 200 & 755\\
Remote & VQA & 5000 & 200 & 37409\\
Remote & Referring & 5000 & 200 & 16159\\
Remote & Captioning & 5000 & 200 & 9350\\
\bottomrule\end{tabular}
\end{suppcompacttable}
\end{table}

\noindent \textbf{Sampling and balancing.}
For surgery, we allocate the remaining sampling budget across labels from rarest to most common and sample without replacement. For underwater imagery, image groups are stratified by source for VQA and referring, and by label for classification. For remote sensing, we sample one annotation per image within each task and draw validation images from the unused training pool. These procedures control supervision volume while retaining the original test distributions. 

\subsection{External and general benchmarks}
External evaluations use the checkpoint selected on supervised source-task validation. PhaKIR contains 5,143 frames sampled at 1~Hz from validation videos 07 and 13. We exclude the undefined phase and map the seven named phases to the corresponding CholecVQA classes. HRVQA uses the 2,000 questions in the HRVQA-2k validation partition. 

\begin{table}[!htbp]
\centering
\setlength{\abovecaptionskip}{3pt}
\setlength{\belowcaptionskip}{3pt}
\footnotesize
\setlength{\tabcolsep}{3pt}
\renewcommand{\arraystretch}{1.0}
\caption{\noindent \textbf{External and general evaluation coverage.}}
\label{tab:supp_benchmarks}
\begin{suppcompacttable}
\begin{tabular}{@{}lll@{}}
\toprule
Benchmark & Evaluation units & Metric\\\midrule
PhaKIR & 3,115 + 2,028 frames & Seven-class macro-F1\\
HRVQA & 2,000 questions & Accuracy\\
MMMU & 150 dev + 900 validation & Accuracy\\
MathVision & 3,040 questions & Accuracy\\
RealWorldQA & 765 questions & Accuracy\\
ODinW & 4,278 images across 12 datasets & Mean COCO AP\\
\bottomrule\end{tabular}
\end{suppcompacttable}
\end{table}

\noindent The ODinW evaluation includes AerialMaritimeDrone large, Aquarium, CottontailRabbits, EgoHands specific, NorthAmericaMushrooms, Packages Raw, PascalVOC, pothole, Raccoon, ShellfishOpenImages, thermalDogsAndPeople, and VehiclesOpenImages. Reported ODinW results therefore average over these 12 datasets.

\section{Prompts}\label{supp:prompts}
\noindent \textbf{Task instructions and method-specific prompts.}
We distinguish three prompting conditions. \emph{Default} requests a constrained output, such as a only the class label or bounding box. \emph{Open} relaxes this restriction, allowing the model to answer openly. \emph{Describe} asks for a visual description of the image before the final answer. SFT uses the open prompt since we do not have expert visual description trajectories, whereas on-policy methods such as GRPO and OPSD use describe-then-answer prompts. 

\noindent \textbf{Base-model sensitivity and prompt choice.}
Table~\ref{tab:supp_prompts} reports the base score under each evaluation recipe. Constrained prompts often reduce base performance, suggesting that they do not consistently elicit the capabilities available before adaptation. Because our annotations provide target answers without intermediate visual reasoning traces, SFT supervises only these reference responses. On-policy methods, however, can learn from student-generated traces using the same available annotations. Specifically, GRPO evaluates sampled responses through task rewards, while OPSD provides teacher feedback along student-generated prefixes conditioned on the reference answer. Restricting generation to a few answer tokens would therefore limit the opportunity to build on the model's existing capabilities. Since we aim to study acquisition and retention under these different forms of supervision, GRPO and OPSD use describe-then-answer prompts, except for captioning, while SFT uses open prompts that do not request reasoning traces absent from its supervision. To account for prompt-dependent starting performance, we compare each adapted model with its prompt-matched base (Eq.~6). 

\begin{table}[!htbp]
\centering
\setlength{\abovecaptionskip}{3pt}
\setlength{\belowcaptionskip}{3pt}
\footnotesize
\setlength{\tabcolsep}{3pt}
\renewcommand{\arraystretch}{1.0}
\caption{\noindent \textbf{Prompting changes the starting point for adaptation.} Qwen3.5-4B scores under default, open, and describe prompts, in each task\textquotesingle s native metric. SFT uses open; GRPO and OPSD use describe.}
\label{tab:supp_prompts}
\begin{suppcompacttable}
\begin{tabular}{@{}lrrr@{}}
\toprule
Domain / task & Default & Open & Describe\\\midrule
Surgery / Binary VQA & 38.11 & 50.36 & 59.73\\
Surgery / Phase & 27.07 & 36.32 & 38.56\\
Surgery / Counting & 44.29 & 48.24 & 50.22\\
Underwater / Classif. & 39.60 & 54.95 & 52.32\\
Underwater / referring & 53.77 & 55.50 & 54.44\\
Underwater / VQA & 55.84 & 61.81 & 57.81\\
Remote / VQA & 67.05 & 69.05 & 67.22\\
Remote / referring & 52.52 & 50.27 & 43.55\\
Remote / captioning & 48.56 & 48.58 & 48.40\\
\bottomrule\end{tabular}
\end{suppcompacttable}
\end{table}

\noindent The describe prompt appends \emph{First, describe what you observe in the image step by step. Then answer the question.} Underwater VQA changes the last sentence to \emph{Then provide a brief answer with a short phrase or single word.} Remote referring prefixes the object description with \emph{Locate this object in the image:} and appends the trace instruction. 

\section{Training details}\label{supp:training}
\subsection{Shared optimization budget}
All methods update the full Qwen3.5-4B model for one epoch, with thinking mode disabled, a learning rate of $2\times10^{-6}$, and batches of 96. Training uses ten warmup steps. The update steps are 62 for surgical task-exclusion runs, 93 for the surgical all-task reference, 104 for task-exclusion runs in the other domains, and 156 for their all-task references. SFT and GRPO use four training H200 GPUs; OPSD assigns three GPUs to the student and one to the teacher. 

\subsection{Method-specific supervision}
SFT optimizes the likelihood of reference response tokens. GRPO and OPSD each sample eight student responses per prompt at temperature 1.0. GRPO uses binary or graded task rewards and centers each reward by the group mean. For tasks with judge-based rewards, GRPO uses NVIDIA-Nemotron-3-Super-120B-A12B-FP8 during training, whereas evaluation uses GPT-OSS-120B (Sec.~\ref{supp:eval}). We separate the training and evaluation judges to reduce the risk that measured gains reflect exploitation of judge-specific scoring behavior rather than improved task performance. OPSD uses a frozen teacher conditioned on the appended reference answer and the sampled student prefix, providing token-level supervision along student-generated trajectories.

\begin{table}[!htbp]
\centering\footnotesize
\caption{\textbf{GRPO reward for each sampled response.}}
\label{tab:supp_rewards}
\begin{tabular}{@{}p{0.33\linewidth}p{0.58\linewidth}@{}}
\toprule Task & Reward rule\\\midrule
Surgical binary VQA, phase, counting & 1 if the extracted yes/no label, phase, or count equals the reference, otherwise 0.\\
Underwater species & 1 for a normalized exact match or an extracted option matching the reference, otherwise 0.\\
Underwater and remote referring & 1 if box IoU is at least 0.5, otherwise 0. LLM extraction is used if parsing fails.\\
Underwater and remote VQA & 1 for an exact match or a positive semantic judgment, otherwise 0. Remote yes/no and integer responses from 0 to 99 use exact matching directly.\\
Remote caption & The judge's CLAIR score divided by 100 and clipped to $[0,1]$.\\
\bottomrule\end{tabular}
\end{table}

\noindent\textbf{OPSD teacher input.} The student samples from the ordinary image and task prompt. After the formatted prompt and before the student's response tokens, the teacher receives the reference annotation from the training example. For a counting example whose reference is 2, the inserted text reads:
\begin{suppprompt}
Reference solution:\\
2\\
Using this as a reference, derive the answer yourself. Think step by step.
\end{suppprompt}
The teacher scores the next token at each prefix of that same sampled response. The reference is visible only to the teacher.

\begin{table}[!htbp]
\centering\footnotesize
\caption{\textbf{Method-specific training settings.} The OPSD entry lists the configured loss settings.}
\label{tab:supp_training}
\begin{suppcompacttable}
\begin{tabular}{@{}llll@{}}
\toprule Setting & SFT & GRPO & OPSD\\\midrule
Response source & Reference & Student & Student\\
Responses per prompt & 1 & 8 & 8\\
Prompt cap & 8,192 (underwater 16,384) & Same & Same\\
Response cap & 2,048 (underwater 512) & Same & Same\\
Policy clipping & -- & Lower 0.2, upper 0.3 & --\\
Supervision & Reference tokens & Task reward & k3 / top-$k$ 32\\
\bottomrule\end{tabular}
\end{suppcompacttable}
\end{table}

\subsection{Checkpoint selection}
Validation runs every five updates for surgery and every ten updates for the other domains. Within each run, the selected checkpoint maximizes the mean online validation score of the supervised tasks.

\section{Evaluation protocol}\label{supp:eval}
\subsection{Answer recovery and task metrics}
Section~\ref{supp:judge_templates} gives the judge and extraction prompts. For surgical binary VQA, phase classification, and counting, we compare recovered class labels with the reference and report macro-F1.

\begin{table}[!htbp]
\centering
\setlength{\abovecaptionskip}{3pt}
\setlength{\belowcaptionskip}{3pt}
\footnotesize
\setlength{\tabcolsep}{3pt}
\renewcommand{\arraystretch}{1.0}
\caption{\noindent \textbf{Task-specific scoring routes.} Referring uses Acc@0.5; class-based surgical tasks and PhaKIR use macro-F1. }
\label{tab:supp_scoring}
\begin{suppcompacttable}
\begin{tabular}{@{}ll@{}}
\toprule
Task & Scoring route\\\midrule
Underwater referring & Parse coordinates, extractor fallback, IoU $\geq$ 0.5\\
Remote referring & Parse coordinates, IoU $\geq$ 0.5\\
Underwater Classif. & Direct class or extracted option, accuracy\\
VQA & Direct match or semantic correctness judgment\\
Caption & Reference-conditioned CLAIR judgment\\
PhaKIR & Extract predicted Phase, macro-F1\\
\bottomrule\end{tabular}
\end{suppcompacttable}
\end{table}

The local evaluator configuration uses GPT-OSS-120B at temperature zero for answer extraction where enabled. Semantic correctness judgments receive the reference answer; referring extraction receives the generated response text. These operations have different roles: extraction recovers a prediction, whereas semantic judging assesses its correctness against a reference.

\subsection{Uncertainty estimation}
The error bars in Fig.~3 of the main paper report the standard deviation across three training runs. For held-out confidence intervals, we pair base and adapted predictions on shared test IDs, draw 3,000 paired cluster-bootstrap replicates with replacement, and recompute the task metric in each replicate. We resample the eight test videos for surgery and individual images for underwater and remote sensing. Models evaluated on the same clusters share the resampling weights. We average the three run-level changes within each replicate and take the 2.5th and 97.5th percentiles. The resulting 95\% intervals are conditional on the observed runs and selected checkpoints. They quantify test-sample uncertainty and should be interpreted separately from variation across training runs.

\section{Complete results}\label{supp:results}
We report absolute scores alongside prompt-matched changes to distinguish the proficiency attained after adaptation from the amount acquired or lost. Each adapted score averages three training runs. \emph{All} denotes the separately trained reference supervised on all three tasks; these rows are excluded from HDR and supervised-selection counts. Bold cells identify the excluded task.

\subsection{Surgery: CholecVQA}
SFT produces both strong positive transfer and substantial degradation within surgery. Holding out binary VQA yields a 22.76-point gain, whereas holding out phase classification yields a 14.20-point loss despite large improvements on both supervised tasks. The rotation therefore changes the conclusion that a fixed held-out task would support.

\begin{table}[!htbp]
\centering
\footnotesize
\setlength{\tabcolsep}{3pt}
\renewcommand{\arraystretch}{1.15}
\caption{\textbf{Surgical acquisition and retention depend on which task is supervised.}
Cells report mean macro-F1 with changes from the prompt-matched base in parentheses.
Bold cells identify performance on the excluded task.
``None'' denotes supervision on all three tasks.}
\label{tab:supp_scores_surgery}
\resizebox{\columnwidth}{!}{%
\begin{tabular}{@{}llrrr@{}}
\toprule
Method & Excluded task & Binary VQA & Counting & Phase \\
\midrule
GRPO & None
& 85.73 (+26.00) & 53.58 (+3.36) & 31.91 (-6.65) \\
& Binary VQA
& \textbf{67.12 (+7.39)} & 54.23 (+4.01) & 27.94 (-10.62) \\
& Counting
& 86.06 (+26.33) & \textbf{51.81 (+1.59)} & 27.19 (-11.37) \\
& Phase
& 86.85 (+27.12) & 54.17 (+3.95) & \textbf{37.64 (-0.92)} \\
\midrule
OPSD & None
& 74.69 (+14.96) & 54.42 (+4.20) & 45.39 (+6.83) \\
& Binary VQA
& \textbf{61.00 (+1.27)} & 54.23 (+4.01) & 45.02 (+6.46) \\
& Counting
& 72.26 (+12.53) & \textbf{48.56 (-1.66)} & 39.87 (+1.31) \\
& Phase
& 76.44 (+16.71) & 55.32 (+5.10) & \textbf{31.94 (-6.62)} \\
\midrule
SFT & None
& 92.82 (+42.46) & 69.05 (+20.81) & 54.12 (+17.80) \\
& Binary VQA
& \textbf{73.12 (+22.76)} & 66.76 (+18.52) & 59.01 (+22.69) \\
& Counting
& 89.60 (+39.24) & \textbf{56.56 (+8.32)} & 60.94 (+24.62) \\
& Phase
& 92.80 (+42.44) & 69.33 (+21.09) & \textbf{22.12 (-14.20)} \\
\bottomrule
\end{tabular}%
}
\end{table}

Table~\ref{tab:supp_ci_surgery} separates variation across training runs from conditional test-sample uncertainty. 

\begin{table}[!tp]
\centering
\setlength{\abovecaptionskip}{3pt}
\setlength{\belowcaptionskip}{3pt}
\footnotesize
\setlength{\tabcolsep}{3pt}
\renewcommand{\arraystretch}{1.0}
\caption{\noindent \textbf{Uncertainty in surgical held-out changes.} SD measures variation across three training runs; lower and upper bounds delimit pointwise 95\% paired cluster-bootstrap intervals conditional on the observed runs and checkpoints.}
\label{tab:supp_ci_surgery}
\begin{suppcompacttable}
\begin{tabular}{@{}llrrrrrr@{}}
\toprule
Method & Excluded & Base & Adapt. & $\Delta$ & SD & Lower & Upper\\\midrule
GRPO & Binary VQA & 59.73 & 67.12 & 7.39 & 5.97 & 6.71 & 8.22\\
GRPO & Counting & 50.22 & 51.81 & 1.59 & 0.95 & -2.30 & 4.87\\
GRPO & Phase & 38.56 & 37.64 & -0.92 & 4.36 & -5.02 & 3.54\\
SFT & Binary VQA & 50.36 & 73.12 & 22.76 & 1.15 & 20.75 & 24.81\\
SFT & Counting & 48.24 & 56.55 & 8.32 & 1.75 & 4.88 & 12.24\\
SFT & Phase & 36.32 & 22.12 & -14.20 & 3.41 & -18.67 & -6.00\\
OPSD & Binary VQA & 59.73 & 61.00 & 1.26 & 0.51 & 0.32 & 2.08\\
OPSD & Counting & 50.22 & 48.55 & -1.67 & 1.23 & -4.75 & 2.10\\
OPSD & Phase & 38.56 & 31.94 & -6.61 & 4.13 & -8.11 & -4.58\\
\bottomrule\end{tabular}
\end{suppcompacttable}
\end{table}

\subsection{Underwater imagery: NAUTILUS}
Underwater results show that substantial degradation does not come with substantial acquisition. With classification excluded, SFT loses 16.87 accuracy points while gaining less than one point on each supervised task. OPSD also loses 10.27 points on excluded VQA. GRPO's held-out mean changes remain close to zero across all three rotations.

\begin{table}[!htbp]
\centering
\footnotesize
\setlength{\tabcolsep}{3pt}
\renewcommand{\arraystretch}{1.15}
\caption{\textbf{Large underwater losses can coexist with little supervised acquisition.}
Cells report mean adapted scores with changes from the prompt-matched base in parentheses. Classification and VQA use accuracy. Referring uses Acc@0.5.
Bold cells identify performance on the excluded task.
``None'' denotes supervision on all three tasks.}
\label{tab:supp_scores_underwater}
\resizebox{\columnwidth}{!}{%
\begin{tabular}{@{}llrrr@{}}
\toprule
Method & Excluded task & Classification & Referring & VQA \\
\midrule
GRPO & None
& 49.02 (-3.30) & 55.89 (+1.45) & 66.60 (+8.79) \\
 & Classification
& \textbf{52.56 (+0.24)} & 56.12 (+1.68) & 67.74 (+9.93) \\
 & Referring
& 54.35 (+2.03) & \textbf{54.26 (-0.18)} & 61.39 (+3.58) \\
 & VQA
& 55.22 (+2.90) & 54.44 (-0.00) & \textbf{57.85 (+0.04)} \\
\midrule
OPSD & None
& 61.89 (+9.57) & 50.95 (-3.49) & 58.58 (+0.77) \\
 & Classification
& \textbf{53.40 (+1.08)} & 53.86 (-0.58) & 59.92 (+2.11) \\
 & Referring
& 65.62 (+13.30) & \textbf{50.95 (-3.49)} & 59.14 (+1.33) \\
 & VQA
& 64.55 (+12.23) & 50.38 (-4.06) & \textbf{47.54 (-10.27)} \\
\midrule
SFT & None
& 80.07 (+25.12) & 56.60 (+1.10) & 61.88 (+0.07) \\
 & Classification
& \textbf{38.08 (-16.87)} & 55.81 (+0.31) & 62.26 (+0.45) \\
 & Referring
& 74.14 (+19.19) & \textbf{54.22 (-1.28)} & 61.56 (-0.25) \\
 & VQA
& 82.09 (+27.14) & 56.91 (+1.41) & \textbf{47.68 (-14.13)} \\
\bottomrule
\end{tabular}%
}
\end{table}

With underwater referring excluded, OPSD increased species classification accuracy from 52.32\% to 65.62\% across three runs (69.09\%, 67.37\%, and 60.40\%).

\begin{table}[!tp]
\centering
\setlength{\abovecaptionskip}{3pt}
\setlength{\belowcaptionskip}{3pt}
\footnotesize
\setlength{\tabcolsep}{3pt}
\renewcommand{\arraystretch}{1.0}
\caption{\noindent \textbf{Uncertainty in underwater held-out changes.} SD reports training-run variation; lower and upper bounds report conditional 95\% paired cluster-bootstrap intervals.}
\label{tab:supp_ci_underwater}
\begin{suppcompacttable}
\begin{tabular}{@{}llrrrrrr@{}}
\toprule
Method & Excluded & Base & Adapt. & $\Delta$ & SD & Lower & Upper\\\midrule
GRPO & Classif. & 52.32 & 52.56 & 0.24 & 0.64 & -1.14 & 1.58\\
GRPO & Referring & 54.44 & 54.26 & -0.18 & 1.11 & -1.55 & 1.24\\
GRPO & VQA & 57.81 & 57.85 & 0.05 & 2.28 & -1.55 & 1.69\\
SFT & Classif. & 54.95 & 38.08 & -16.87 & 1.05 & -20.98 & -12.63\\
SFT & Referring & 55.50 & 54.22 & -1.28 & 1.22 & -2.87 & 0.31\\
SFT & VQA & 61.81 & 47.68 & -14.14 & 0.55 & -16.55 & -11.72\\
OPSD & Classif. & 52.32 & 53.40 & 1.08 & 1.29 & -1.21 & 3.33\\
OPSD & Referring & 54.44 & 50.95 & -3.49 & 0.33 & -5.21 & -1.85\\
OPSD & VQA & 57.81 & 47.54 & -10.27 & 5.61 & -12.56 & -8.09\\
\bottomrule\end{tabular}
\end{suppcompacttable}
\end{table}

\subsection{Remote sensing: VRSBench}
All three methods improve excluded captioning, but this positive transfer does not extend uniformly to the other tasks. SFT loses performance on excluded referring, while OPSD loses performance on excluded VQA. For referring, the run-level standard deviations also show that near-zero mean changes can coexist with substantial variation across training repeats.

\begin{table}[!htbp]
\centering
\footnotesize
\setlength{\tabcolsep}{3pt}
\renewcommand{\arraystretch}{1.15}
\caption{\textbf{Positive transfer to remote-sensing captioning does not imply uniform retention.}
Cells report mean adapted scores with changes from the prompt-matched base in parentheses. Captioning uses CLAIR, referring uses Acc@0.5, and VQA uses accuracy.
Bold cells identify performance on the excluded task.
``None'' denotes supervision on all three tasks.}
\label{tab:supp_scores_remote}
\resizebox{\columnwidth}{!}{%
\begin{tabular}{@{}llrrr@{}}
\toprule
Method & Excluded task & Caption & Referring & VQA \\
\midrule
GRPO & None
& 69.58 (+21.18) & 62.38 (+18.83) & 75.65 (+8.43) \\
 & Caption
& \textbf{51.19 (+2.79)} & 64.51 (+20.96) & 73.50 (+6.28) \\
 & Referring
& 65.43 (+17.03) & \textbf{43.48 (-0.07)} & 75.30 (+8.08) \\
 & VQA
& 50.01 (+1.61) & 64.62 (+21.07) & \textbf{69.75 (+2.53)} \\
\midrule
OPSD & None
& 48.34 (-0.06) & 54.02 (+10.47) & 70.64 (+3.42) \\
 & Caption
& \textbf{54.01 (+5.61)} & 54.34 (+10.79) & 70.49 (+3.27) \\
 & Referring
& 48.87 (+0.47) & \textbf{43.58 (+0.03)} & 70.82 (+3.60) \\
 & VQA
& 47.48 (-0.92) & 51.23 (+7.68) & \textbf{66.64 (-0.58)} \\
\midrule
SFT & None
& 57.42 (+8.84) & 71.10 (+20.83) & 75.95 (+6.90) \\
 & Caption
& \textbf{53.87 (+5.29)} & 71.92 (+21.65) & 74.30 (+5.25) \\
 & Referring
& 57.29 (+8.71) & \textbf{48.92 (-1.35)} & 76.23 (+7.18) \\
 & VQA
& 56.37 (+7.79) & 70.91 (+20.64) & \textbf{74.79 (+5.74)} \\
\bottomrule
\end{tabular}%
}
\end{table}

\begin{table}[!tp]
\centering
\setlength{\abovecaptionskip}{3pt}
\setlength{\belowcaptionskip}{3pt}
\footnotesize
\setlength{\tabcolsep}{3pt}
\renewcommand{\arraystretch}{1.0}
\caption{\noindent \textbf{Uncertainty in remote-sensing held-out changes.} Training-run SD and conditional 95\% paired cluster-bootstrap bounds describe different sources of variation and should be read separately.}
\label{tab:supp_ci_remote}
\begin{suppcompacttable}
\begin{tabular}{@{}llrrrrrr@{}}
\toprule
Method & Excluded & Base & Adapt. & $\Delta$ & SD & Lower & Upper\\\midrule
GRPO & Captioning & 48.40 & 51.19 & 2.79 & 4.06 & 2.30 & 3.28\\
GRPO & Referring & 43.55 & 43.48 & -0.07 & 15.99 & -0.69 & 0.56\\
GRPO & VQA & 67.22 & 69.75 & 2.53 & 2.83 & 2.14 & 2.94\\
SFT & Captioning & 48.58 & 53.87 & 5.29 & 1.75 & 4.77 & 5.86\\
SFT & Referring & 50.27 & 48.92 & -1.35 & 9.09 & -1.90 & -0.82\\
SFT & VQA & 69.05 & 74.79 & 5.75 & 0.29 & 5.30 & 6.18\\
OPSD & Captioning & 48.40 & 54.01 & 5.61 & 0.55 & 5.15 & 6.09\\
OPSD & Referring & 43.55 & 43.58 & 0.03 & 2.01 & -0.52 & 0.62\\
OPSD & VQA & 67.22 & 66.64 & -0.58 & 2.60 & -0.95 & -0.22\\
\bottomrule\end{tabular}
\end{suppcompacttable}
\end{table}

\subsection{Cost of supervised selection}
For each task-exclusion split, we select the method with the highest mean supervised validation score, $V_s(m)$. The selection cost is the magnitude of its negative held-out change, or zero if the change is nonnegative. Table~\ref{tab:supp_selection} reports the selected method and its absolute held-out performance. Selection produces a negative held-out mean in four of nine splits, including a 14.20-point loss on surgical phase classification and a 14.14-point loss on underwater VQA. The underwater referring cost is 3.49 points.

\begin{table}[!htbp]
\centering
\setlength{\abovecaptionskip}{3pt}
\setlength{\belowcaptionskip}{3pt}
\footnotesize
\setlength{\tabcolsep}{3pt}
\renewcommand{\arraystretch}{1.0}
\caption{\noindent \textbf{Supervised validation can select configurations that degrade excluded capabilities.} The selected method maximizes $V_s(m)$. Cost is the negative held-out change truncated below at zero, in the held-out task\textquotesingle s native metric points.}
\label{tab:supp_selection}
\begin{suppcompacttable}
\begin{tabular}{@{}lllrrrr@{}}
\toprule
Domain & Excluded & Method & V & Adapt. & Base & Cost\\\midrule
Surgery & VQA & SFT & 0.72 & 73.12 & 50.36 & 0.00\\
Surgery & Counting & SFT & 0.78 & 56.55 & 48.24 & 0.00\\
Surgery & Phase & SFT & 0.88 & 22.12 & 36.32 & 14.20\\
Underwater & Classif. & GRPO & 0.59 & 52.56 & 52.32 & 0.00\\
Underwater & Refer. & OPSD & 0.57 & 50.95 & 54.44 & 3.49\\
Underwater & VQA & SFT & 0.48 & 47.68 & 61.81 & 14.14\\
Remote & Caption & GRPO & 0.70 & 51.19 & 48.40 & 0.00\\
Remote & Refer. & GRPO & 0.81 & 43.48 & 43.55 & 0.07\\
Remote & VQA & GRPO & 0.80 & 69.75 & 67.22 & 0.00\\
\bottomrule\end{tabular}
\end{suppcompacttable}
\end{table}

\subsection{Sensitivity to degradation tolerance}
HDR counts splits whose mean held-out change falls strictly below the negative tolerance. Table~\ref{tab:supp_tolerance} reports the corresponding counts across nine splits per method, using unrounded native-metric differences. The same numerical tolerance is applied to each task's native metric; it does not imply that a one-point change has equivalent practical significance across tasks.

\begin{table}[!htbp]
\centering
\setlength{\abovecaptionskip}{3pt}
\setlength{\belowcaptionskip}{3pt}
\footnotesize
\setlength{\tabcolsep}{3pt}
\renewcommand{\arraystretch}{1.0}
\caption{\noindent \textbf{GRPO\textquotesingle s negative held-out means are concentrated near zero.} Entries count splits with mean change strictly below $-\epsilon$, out of nine per method. Tolerances use native metric points, not significance thresholds.}
\label{tab:supp_tolerance}
\begin{suppcompacttable}
\begin{tabular}{@{}lrrr@{}}
\toprule
Tolerance & GRPO & OPSD & SFT\\\midrule
0.00 & 3.00 & 5.00 & 5.00\\
0.10 & 2.00 & 5.00 & 5.00\\
0.50 & 1.00 & 5.00 & 5.00\\
1.00 & 0.00 & 4.00 & 5.00\\
2.00 & 0.00 & 3.00 & 3.00\\
5.00 & 0.00 & 2.00 & 3.00\\
\bottomrule\end{tabular}
\end{suppcompacttable}
\end{table}

At a tolerance of one native metric point, degradation counts remain 5/9 for SFT and fall to 4/9 for OPSD and 0/9 for GRPO. Thus, GRPO's negative means are concentrated near zero, whereas several losses under SFT and OPSD persist at larger tolerances. These counts describe magnitude rather than statistical significance.

\section{Additional analyses}\label{supp:analyses}
\subsection{GRPO on supervised surgical phase classification}\label{supp:grpo_phase}
GRPO's limited degradation on excluded tasks coexists with a failure to acquire surgical phase classification when it is supervised. To characterize this behavior, we examine phase-classification rollout groups up to the selected checkpoints. When binary VQA is excluded, 2,503 of 6,961 groups (36.0\%) contain eight zero rewards. With counting excluded, this occurs in 2,744 of 5,797 groups (47.3\%). Including all-correct groups raises the proportions with zero centered advantage to 49.1\% and 65.3\%, respectively. Such groups provide no within-group reward contrast under the stated GRPO objective.

For clipping/cutting, all eight responses are incorrect in 86.2\% and 89.2\% of groups in the two settings. These proportions rise to 93.9\% and 98.4\% in the final ten steps before checkpoint selection. The corresponding class F1 falls from 19.35 to 1.08 and 0.00 (Table~\ref{tab:supp_grpo_phase}).

\begin{table}[!htbp]
\centering
\setlength{\abovecaptionskip}{3pt}
\setlength{\belowcaptionskip}{3pt}
\footnotesize
\setlength{\tabcolsep}{3pt}
\renewcommand{\arraystretch}{1.0}
\caption{\noindent \textbf{GRPO can lose supervised phase-recognition capability.} Per-class F1 is shown against the matched base, alongside training-class counts (GB: gallbladder). Phase classification is supervised in both task-exclusion settings.}
\label{tab:supp_grpo_phase}
\begin{suppcompacttable}
\begin{tabular}{@{}lrrrr@{}}
\toprule
Phase & Train $N$ & Base F1 & \shortstack{GRPO\\excl. VQA} & \shortstack{GRPO\\excl. count}\\\midrule
Calot dissection & 933 & 77.14 & 67.89 & 55.38\\
Cleaning/coag. & 391 & 15.19 & 21.15 & 9.80\\
Clipping/cutting & 275 & 19.35 & 1.08 & 0.00\\
GB dissection & 933 & 63.16 & 32.13 & 31.95\\
GB packaging & 162 & 55.07 & 68.32 & 60.99\\
GB retraction & 137 & 19.29 & 0.00 & 0.00\\
Preparation & 169 & 20.69 & 5.03 & 32.22\\
\bottomrule\end{tabular}
\end{suppcompacttable}
\end{table}

The two dissection phases account for 62.2\% of training frames and 78.4\% and 76.2\% of groups with mixed rewards, respectively. 

\subsection{SFT transfer with surgical phase classification excluded}\label{supp:surgery_transfer}
All 3,000 binary VQA training questions name a phase using 49 templates, but supervise only a yes/no response. Counting questions do not name a phase. Thus, excluding phase classification removes direct supervision of the image-to-phase mapping while retaining phase names in the binary VQA inputs.

Across three SFT runs, all 3,132 held-out phase responses are canonical class names. Nevertheless, gallbladder dissection receives 84.0\% of predictions while accounting for only 16.4\% of test targets (Table~\ref{tab:supp_sft_phase}). The most frequently named phase in binary VQA training is Calot triangle dissection, so the dominant predicted class is not simply the most frequently mentioned phase.

\begin{table}[!htbp]
\centering
\setlength{\abovecaptionskip}{3pt}
\setlength{\belowcaptionskip}{3pt}
\footnotesize
\setlength{\tabcolsep}{3pt}
\renewcommand{\arraystretch}{1.0}
\caption{\noindent \textbf{Valid phase labels do not establish retained phase recognition.} With phase classification excluded, SFT assigns 84.0\% of predictions to gallbladder dissection. GB denotes gallbladder. Test counts and target shares describe the same test set; prediction shares pool three runs.}
\label{tab:supp_sft_phase}
\begin{suppcompacttable}
\begin{tabular}{@{}lrrrrr@{}}
\toprule
Phase & Test $N$ & GT \% & Pred. \% & Base F1 & SFT F1\\\midrule
Calot dissection & 557 & 53.35 & 5.52 & 77.65 & 12.68\\
Cleaning/coag. & 56 & 5.36 & 0.51 & 9.68 & 2.09\\
Clipping/cutting & 117 & 11.21 & 0.06 & 8.96 & 0.56\\
GB dissection & 171 & 16.38 & 84.00 & 33.82 & 29.65\\
GB packaging & 44 & 4.21 & 3.07 & 72.22 & 74.54\\
GB retraction & 57 & 5.46 & 0.99 & 47.54 & 12.32\\
Preparation & 42 & 4.02 & 5.84 & 4.35 & 22.97\\
\bottomrule\end{tabular}
\end{suppcompacttable}
\end{table}

The prompt-matched base achieves 36.32 macro-F1, compared with 22.12 after adaptation. Valid phase names therefore do not establish retained phase-recognition capability: the loss occurs in assigning phases to images despite valid output labels.

\subsection{Transfer to external in-domain datasets}\label{supp:hrvqa}\label{supp:external}
\noindent \textbf{PhaKIR: dependence on direct phase supervision persists.}
All external evaluations reuse the source-selected checkpoints. On PhaKIR, SFT improves phase classification when that task receives source supervision, including under all-task training, but loses 8.71 macro-F1 points when phase classification is excluded. GRPO loses performance in all four settings. These results show that the source-dataset pattern can extend to a different surgical dataset.

\begin{table}[!htbp]
\centering
\setlength{\abovecaptionskip}{3pt}
\setlength{\belowcaptionskip}{3pt}
\footnotesize
\setlength{\tabcolsep}{3pt}
\renewcommand{\arraystretch}{1.0}
\caption{\noindent \textbf{SFT\textquotesingle s dependence on phase supervision extends to PhaKIR.} Scores are seven-class macro-F1 on a 0--100 scale; changes are macro-F1 points relative to the prompt-matched base. All denotes supervision on all three source tasks.}
\label{tab:supp_phakir}
\begin{suppcompacttable}
\begin{tabular}{@{}llrrrr@{}}
\toprule
Excluded & Method & Base & Adapt. & $\Delta$ & Runs\\\midrule
All & GRPO & 29.05 & 27.86 & -1.19 & 3\\
All & OPSD & 29.05 & 33.27 & 4.22 & 3\\
All & SFT & 31.98 & 41.22 & 9.24 & 3\\
VQA & GRPO & 29.05 & 25.50 & -3.55 & 3\\
VQA & OPSD & 29.05 & 34.71 & 5.66 & 3\\
VQA & SFT & 31.98 & 44.12 & 12.14 & 3\\
Counting & GRPO & 29.05 & 26.37 & -2.68 & 3\\
Counting & OPSD & 29.05 & 31.27 & 2.22 & 3\\
Counting & SFT & 31.98 & 42.74 & 10.76 & 3\\
Phase & GRPO & 29.05 & 28.45 & -0.60 & 3\\
Phase & OPSD & 29.05 & 28.46 & -0.59 & 3\\
Phase & SFT & 31.98 & 23.27 & -8.71 & 3\\
\bottomrule\end{tabular}
\end{suppcompacttable}
\end{table}

\noindent \textbf{HRVQA: source-task acquisition does not ensure external transfer.}
All twelve remote-sensing configurations lose accuracy on HRVQA, including configurations that improve VRSBench VQA. The losses span 10.68--24.22 accuracy points (Table~\ref{tab:supp_hrvqa}). This external evaluation therefore reveals a limitation that source-task scores alone do not establish.

\begin{table}[!htbp]
\centering
\setlength{\abovecaptionskip}{3pt}
\setlength{\belowcaptionskip}{3pt}
\footnotesize
\setlength{\tabcolsep}{3pt}
\renewcommand{\arraystretch}{1.0}
\caption{\noindent \textbf{Every remote-sensing configuration loses accuracy on HRVQA.} Scores are accuracy on a 0--100 scale; changes are accuracy points relative to the prompt-matched base. All denotes supervision on all three source tasks.}
\label{tab:supp_hrvqa}
\begin{suppcompacttable}
\begin{tabular}{@{}llrrrr@{}}
\toprule
Excluded & Method & Base & Adapt. & $\Delta$ & Runs\\\midrule
All & GRPO & 48.85 & 33.87 & -14.98 & 3\\
All & OPSD & 48.85 & 36.25 & -12.60 & 3\\
All & SFT & 47.25 & 28.30 & -18.95 & 3\\
Caption & GRPO & 48.85 & 31.23 & -17.62 & 3\\
Caption & OPSD & 48.85 & 35.18 & -13.67 & 3\\
Caption & SFT & 47.25 & 28.15 & -19.10 & 3\\
Refer. & GRPO & 48.85 & 31.28 & -17.57 & 3\\
Refer. & OPSD & 48.85 & 35.07 & -13.78 & 3\\
Refer. & SFT & 47.25 & 23.03 & -24.22 & 3\\
VQA & GRPO & 48.85 & 37.12 & -11.73 & 3\\
VQA & OPSD & 48.85 & 38.17 & -10.68 & 3\\
VQA & SFT & 47.25 & 36.03 & -11.22 & 3\\
\bottomrule\end{tabular}
\end{suppcompacttable}
\end{table}

\noindent \textbf{Which questions contribute to the HRVQA decline?}
We pair base and adapted predictions on each of the 2,000 questions and decompose the net accuracy loss by reference-answer category. For category $c$ and run $r$, its contribution is
\begin{equation}
L_{c,r}=\frac{100}{2{,}000}\left(N^{\mathrm{correct}\to\mathrm{wrong}}_{c,r}-N^{\mathrm{wrong}\to\mathrm{correct}}_{c,r}\right).
\end{equation}
We average these contributions across the three runs per configuration. Positive values denote losses in overall accuracy points; negative values denote net gains. The final column of Table~\ref{tab:supp_hrvqa_categories} expresses the explicit-absence contribution as a percentage of the total net decline. Contributions use the full test-set denominator, rather than category-specific accuracy denominators.

\begin{table}[!htbp]
\centering
\setlength{\abovecaptionskip}{3pt}
\setlength{\belowcaptionskip}{3pt}
\footnotesize
\setlength{\tabcolsep}{3pt}
\renewcommand{\arraystretch}{1.0}
\caption{\noindent \textbf{Negative-reference questions contribute most of the net HRVQA decline.} Contributions are overall accuracy-point losses; negative entries denote net gains. Absence share is the explicit-absence contribution divided by total net loss. Category totals may differ by rounding.}
\label{tab:supp_hrvqa_categories}
\begin{suppcompacttable}
\begin{tabular}{@{}llrrrrr@{}}
\toprule
Method & Excluded & Absence & \shortstack{Other\\neg.} & Positive & Total & \shortstack{Absence\\share (\%)}\\\midrule
GRPO & All & 5.67 & 9.88 & -0.57 & 14.98 & 37.82\\
GRPO & Caption & 6.75 & 12.95 & -2.08 & 17.62 & 38.32\\
GRPO & Refer. & 7.60 & 13.48 & -3.52 & 17.57 & 43.26\\
GRPO & VQA & 3.08 & 5.93 & 2.72 & 11.73 & 26.28\\
OPSD & All & 2.90 & 7.78 & 1.92 & 12.60 & 23.02\\
OPSD & Caption & 3.10 & 7.95 & 2.62 & 13.67 & 22.68\\
OPSD & Refer. & 3.53 & 7.87 & 2.38 & 13.78 & 25.63\\
OPSD & VQA & 2.37 & 5.42 & 2.90 & 10.68 & 22.15\\
SFT & All & 4.82 & 9.47 & 4.67 & 18.95 & 25.42\\
SFT & Caption & 6.17 & 11.57 & 1.37 & 19.10 & 32.29\\
SFT & Refer. & 5.62 & 10.92 & 7.68 & 24.22 & 23.19\\
SFT & VQA & 4.27 & 4.78 & 2.17 & 11.22 & 38.04\\
\bottomrule\end{tabular}
\end{suppcompacttable}
\end{table}

Explicit absence accounts for 248 questions (12.4\% of the test set). The other negative-reference categories contain 25 negative Boolean responses, 170 zero counts, and 254 unanswerable cases; the remaining 1,303 questions have positive references. Explicit absence contributes 22.2--43.3\% of the net decline across the twelve configurations. The broader negative-reference group contributes a majority in every configuration.

These categories describe question and reference semantics, and training repeats reuse the same questions. The decomposition supports a concentration of losses on negative-reference questions.

\subsection{Agreement between general and in-domain retention}\label{supp:general}
Each of the 27 method--split configurations contributes one held-out change and one change per general benchmark. We count a disagreement when the general-benchmark change is nonnegative but the held-out change is negative. The conditional bootstrap ranges resample held-out test units while keeping general-benchmark scores and selected checkpoints fixed. They therefore do not incorporate uncertainty in the general-benchmark evaluations.

\begin{table}[!htbp]
\centering
\setlength{\abovecaptionskip}{3pt}
\setlength{\belowcaptionskip}{3pt}
\footnotesize
\setlength{\tabcolsep}{3pt}
\renewcommand{\arraystretch}{1.0}
\caption{\noindent \textbf{General retention can coexist with excluded-task degradation.} Disagreements have nonnegative general change and negative held-out change. For the three benchmarks shown, bounds are the 2.5th and 97.5th percentiles of disagreement counts with general scores and checkpoints fixed.}
\label{tab:supp_general_agreement}
\begin{suppcompacttable}
\begin{tabular}{@{}lrrrr@{}}
\toprule
Benchmark & Configs. & Disagree. & Lower & Upper\\\midrule
MathVision & 27 & 2 & 0.00 & 3.00\\
ODinW & 27 & 2 & 0.00 & 4.00\\
RealWorldQA & 27 & 10 & 7.00 & 13.00\\
\bottomrule\end{tabular}
\end{suppcompacttable}
\end{table}

\begin{figure}[!tp]
\centering
\includegraphics[width=\linewidth]{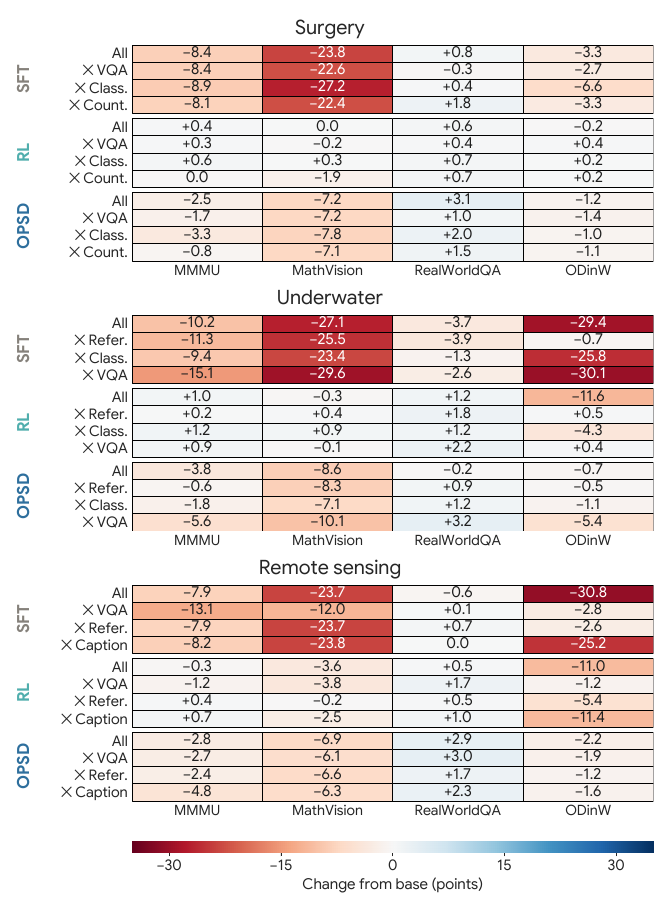}
\caption{\noindent \textbf{General-retention profiles vary across benchmarks and adaptation strategies.} Changes relative to the base model are shown by domain, method, and supervision setting, including all-task references. Values are accuracy-point changes for MMMU, MathVision, and RealWorldQA, and mean COCO AP-point changes for ODinW. These evaluations complement the excluded-task results; preserving a general benchmark does not establish retention within the target domain.}
\label{fig:supp_general}
\end{figure}

General-benchmark prompts are shared across methods. The correlations in Fig.~5 of the main paper use nine task-exclusion configurations per domain and describe variation over those configurations. Figure~\ref{fig:supp_general} additionally includes the all-task references.

\section{Surgical transfer from a weaker starting model}\label{supp:arch}
We additionally apply rotating task exclusion to Gemma-3-4B-IT on CholecVQA. Gemma starts below Qwen3.5-4B on binary VQA, counting, and phase classification under both prompts. This comparison examines whether supervision on two surgical tasks can improve an excluded capability from a weaker starting point. Each adapted model is compared with its own prompt-matched Gemma base.

Table~\ref{tab:supp_gemma_surgery} covers all nine method--split combinations in surgery, alongside the all-task references. Held-out macro-F1 increased in eight of the nine selected task-exclusion runs, with gains above one point in seven.

\noindent \textbf{Positive transfer from a weaker base.}
SFT on phase classification and counting raises excluded binary VQA from 27.24 to 39.91 macro-F1 (+12.67). GRPO on binary VQA and phase classification raises excluded counting from 19.42 to 29.04 macro-F1 (+9.62). These results show that supervision on neighboring surgical tasks can improve capabilities excluded from adaptation.

\begin{table}[!htbp]
\centering
\footnotesize
\setlength{\tabcolsep}{3pt}
\renewcommand{\arraystretch}{1.15}
\caption{\textbf{Surgical task acquisition and held-out transfer with Gemma.}
Each row reports macro-F1 for one randomly selected completed run, with changes from the prompt-matched Gemma base in parentheses.
Bold cells identify performance on the excluded task.
``None'' denotes supervision on all three tasks.}
\label{tab:supp_gemma_surgery}
\resizebox{\columnwidth}{!}{%
\begin{tabular}{@{}llrrr@{}}
\toprule
Method & Excluded task & Binary VQA & Counting & Phase \\
\midrule
GRPO & None
& 48.50 (-0.45) & 25.42 (+6.00) & 12.91 (+2.22) \\
 & Binary VQA
& \textbf{54.74 (+5.79)} & 28.10 (+8.68) & 13.57 (+2.88) \\
 & Counting
& 38.79 (-10.16) & \textbf{29.04 (+9.62)} & 12.96 (+2.27) \\
 & Phase
& 51.69 (+2.74) & 30.07 (+10.65) & \textbf{12.67 (+1.98)} \\
\midrule
OPSD & None
& 64.00 (+15.05) & 22.67 (+3.25) & 11.96 (+1.27) \\
 & Binary VQA
& \textbf{52.12 (+3.17)} & 21.45 (+2.03) & 12.61 (+1.92) \\
 & Counting
& 63.96 (+15.01) & \textbf{19.40 (-0.02)} & 10.91 (+0.22) \\
 & Phase
& 62.96 (+14.01) & 23.80 (+4.38) & \textbf{12.75 (+2.06)} \\
\midrule
SFT & None
& 83.81 (+56.57) & 45.31 (+25.78) & 27.24 (+16.77) \\
 & Binary VQA
& \textbf{39.91 (+12.67)} & 45.53 (+26.00) & 29.35 (+18.88) \\
 & Counting
& 83.21 (+55.97) & \textbf{19.62 (+0.09)} & 29.55 (+19.08) \\
 & Phase
& 81.87 (+54.63) & 45.74 (+26.21) & \textbf{11.49 (+1.02)} \\
\bottomrule
\end{tabular}%
}
\end{table}

\noindent \textbf{Positive transfer does not establish strong task performance.}
With phase classification excluded under SFT, Gemma moves from 10.47 to 11.49 macro-F1. For the same exclusion, Qwen falls from 36.32 to 22.12 macro-F1 on average across three runs, yet still finishes above the selected Gemma run. Thus, a small change from a weak baseline and a substantial loss from a stronger baseline can lead to different conclusions about retention and final proficiency. We therefore report starting capability, absolute adapted performance, and held-out change separately. This surgical comparison does not establish that a weaker base model is generally more resistant to degradation.

\section{Judge and extraction prompts}\label{supp:judge_templates}

\subsection{Surgery}

\par\addvspace{3pt}\noindent{\small CholecVQA, original semantic judgment}\par\nobreak
\begin{suppprompt}
Question: \{question\}\\
Ground Truth Answer: \{ground\_truth\}\\
Predicted Answer: \{predicted\}\\
Does the predicted answer match the ground truth? Answer 1 for match and 0 for not match. Use semantic meaning not exact match.
\end{suppprompt}

\par\addvspace{3pt}\noindent{\small Phase extraction (PhaKIR and CholecVQA)}\par\nobreak
\begin{suppprompt}
Text: \{predicted\}\\
Extract the surgical phase mentioned in the above text. Reply with exactly one of the following options (use lowercase, no punctuation):\\
preparation \textbar{} calot triangle dissection \textbar{} clipping cutting \textbar{} gallbladder dissection \textbar{} cleaning coagulation \textbar{} gallbladder packaging \textbar{} gallbladder retraction\\
Output only the phase name, nothing else.
\end{suppprompt}
\subsection{Underwater}

\par\addvspace{3pt}\noindent{\small NAUTILUS, VQA}\par\nobreak
\begin{suppprompt}
Question: \{question\}\\
Ground Truth Answer: \{ground\_truth\}\\
Predicted Answer: \{predicted\}\\
Does the predicted answer correctly answer the question? Answer 1 for correct and 0 for incorrect. Use semantic meaning, not exact match. Synonyms count as correct.
\end{suppprompt}

\par\addvspace{3pt}\noindent{\small NAUTILUS, species extraction}\par\nobreak
\begin{suppprompt}
Original question: \{question\}\\
Model answer: \{predicted\}\\
The question asks to classify a fish species from a provided list. Extract the final predicted class name from the model answer. Output only the class name exactly as it appears in the question's option list, nothing else.
\end{suppprompt}

\par\addvspace{3pt}\noindent{\small NAUTILUS, box extraction}\par\nobreak
\begin{suppprompt}
Text: \{predicted\}\\
Extract the final bounding box from the above text. Output only the JSON object in the format \{"bbox\_2d": [x1, y1, x2, y2]\} with integer coordinates from 0 to 1000, nothing else.
\end{suppprompt}
\subsection{Remote sensing}

\par\addvspace{3pt}\noindent{\small VRSBench and HRVQA, VQA}\par\nobreak
\begin{suppprompt}
Question: \{question\}, Ground Truth Answer: \{ground\_truth\}, Predicted Answer: \{predicted\}. Does the predicted answer match the ground truth? Answer 1 for match and 0 for not match. Use semantic meaning not exact match. Synonyms are also treated as a match.
\end{suppprompt}

\par\addvspace{3pt}\noindent{\small HRVQA, area questions}\par\nobreak
\begin{suppprompt}
Question: \{question\}, Ground Truth Answer: \{ground\_truth\}, Predicted Answer: \{predicted\}. Does the predicted answer match the ground truth? Answer 1 for match and 0 for not match. The ground truth is a numeric area value. Consider the answer correct if the predicted numeric value is within 10\% of the ground truth (ignore units, compare numbers only).
\end{suppprompt}

\par\addvspace{3pt}\noindent{\small VRSBench, caption similarity}\par\nobreak
\begin{suppprompt}
You are trying to tell if a candidate caption is describing the same image as a reference caption.\\
Candidate: \{candidate\}\\
Reference: \{reference\}\\
On a precise scale from 0 to 100, how likely is it that the candidate describes the same image as the reference? Respond with JSON: \{"score": \textless{}int 0-100\textgreater{}\}
\end{suppprompt}

\subsection{Training judge prompts}
GRPO uses the NAUTILUS templates above for the corresponding tasks and the VRSBench caption template for caption rewards. For remote VQA, the training prompt expands the final sentence to \emph{Synonyms are also treated as a match, e.g., pond and swimming pool.} The remaining training extractors use the following prompts.

\par\addvspace{3pt}\noindent{\small Surgical binary VQA}\par\nobreak
\begin{suppprompt}
Text: \{predicted\}\\
What is the yes or no answer given in the above text? If the text states a positive/affirmative answer, output "yes". If the text states a negative answer, output "no". Output only the single word "yes" or "no", nothing else.
\end{suppprompt}

\par\addvspace{3pt}\noindent{\small Surgical counting}\par\nobreak
\begin{suppprompt}
Text: \{predicted\}\\
Extract the number of tools stated in the above text. Output only a single digit (0, 1, 2, 3, or 4), nothing else.
\end{suppprompt}

\par\addvspace{3pt}\noindent{\small Remote referring}\par\nobreak
\begin{suppprompt}
Text: \{predicted\}\\
Extract the final bounding box answer from the above text. Output only the bounding box in the format [x1,y1,x2,y2] with integer coordinates (values 0 to 1000), nothing else.
\end{suppprompt}

\end{document}